%% file: bare_jrnl.tex
\documentclass[journal]{IEEEtran}

\usepackage[amssymb]{SIunits}
\usepackage{graphicx}
\usepackage{amssymb}
\usepackage{amsthm}
\usepackage{multirow}
\usepackage{enumerate}
\usepackage{enumitem}
\usepackage{wrapfig}
\usepackage{seqsplit}

\usepackage{amsmath}
\usepackage[cmintegrals]{newtxmath}
\usepackage{bm}
\usepackage{footnote}
\usepackage{threeparttable}
\usepackage{epsfig}
\usepackage{graphicx}
\usepackage{amsmath}
\usepackage{amssymb}
\usepackage{balance}
\usepackage{mathtools}
\usepackage{placeins}
\usepackage{pifont}%
\usepackage[misc]{ifsym}
\usepackage[super]{nth}

\usepackage{url}
\usepackage{booktabs}
\usepackage{amsfonts}
\usepackage{nicefrac}
\usepackage{microtype}
\usepackage{lipsum}
\usepackage[hidelinks,colorlinks=true,urlcolor=blue,citecolor=blue,allcolors=blue]{hyperref}
\usepackage{graphicx}
\usepackage{color}
\usepackage[table,dvipsnames]{xcolor}
\usepackage{todonotes}
\usepackage{marginnote}
\presetkeys{todonotes}{size=\tiny}{}

\usepackage{array}
\usepackage{capt-of}
\newcommand{\PreserveBackslash}[1]{\let\temp=\\#1\let\\=\temp}
\newcolumntype{C}[1]{>{\PreserveBackslash\centering}p{#1}}
\newcolumntype{R}[1]{>{\PreserveBackslash\raggedleft}p{#1}}
\newcolumntype{L}[1]{>{\PreserveBackslash\raggedright}p{#1}}

\newcommand{\rom}[1]{\uppercase\expandafter{\romannumeral #1\relax}}

\newcommand{\ie}{{i.e.},~}
\newcommand{\eg}{{e.g.},~}

\newcommand{\xx}{\ding{53}}
\newcommand{\cc}{\checkmark}

\graphicspath{%
    {./figures/}%
}

\ifCLASSOPTIONcompsoc
  \usepackage[caption=false,font=footnotesize,labelfont=sf,textfont=sf]{subfig}
\else
  \usepackage[caption=false,font=footnotesize]{subfig}
\fi

\ifCLASSOPTIONcompsoc
  \usepackage[nocompress]{cite}
\else
  \usepackage{cite}
\fi

\ifCLASSINFOpdf
\else
\fi

\usepackage{algorithm}
\usepackage{algorithmic}

\ifCLASSOPTIONcompsoc
 \usepackage[caption=false,font=normalsize,labelfont=sf,textfont=sf]{subfig}
\else
 \usepackage[caption=false,font=footnotesize]{subfig}
\fi

\usepackage{url}

\begin{document}
\title{AirLeg: A Robust and Accurate Self-Supervised Visual-Inertial-Leg SLAM System}

\author{Kuan~Xu$^{1}$, Tianxin~Hu$^{1}$, Ruofei~Bai$^{1,2}$, Ruimeng~Liu$^{1}$, Yuefan~Hao$^{3}$, Zeyu~Jiang$^{1}$, Zitong~Zhan$^{3}$,\\
        Shenghai~Yuan$^{1}$,~\IEEEmembership{Member,~IEEE,}
        Chen~Wang$^{3}$,~\IEEEmembership{Senior Member,~IEEE,}
        Lihua~Xie$^{4}$,~\IEEEmembership{Fellow,~IEEE}\\
        {\normalsize Project page: \url{https://sairlab.org/AirLeg_homepage/}}%
\thanks{$^{1}$Kuan Xu, Tianxin Hu, Ruofei Bai, Ruimeng Liu, Zeyu Jiang, and Shenghai Yuan are with the Centre for Advanced Robotics Technology Innovation (CARTIN), School of Electrical and Electronic Engineering, Nanyang Technological University, 50 Nanyang Avenue, Singapore 639798, {\tt \{kuan.xu, tianxin.hu, ruimeng.liu, shyuan\}@ntu.edu.sg}, {\tt zjiang015@e.ntu.edu.sg}.}%
\thanks{$^{2}$Ruofei Bai is also with the A*STAR Institute of Advanced Intelligence and Computing (A*STAR IAIC), Singapore, {\tt ruofei001@e.ntu.edu.sg}.}%
\thanks{$^{3}$Yuefan Hao, Zitong Zhan, and Chen Wang are with the Spatial AI \& Robotics Lab, Department of Computer Science and Engineering, University at Buffalo, Buffalo, NY 14260, USA, {\tt yuefan.hao@outlook.com}, {\tt \{zitongz, chenw\}@sairlab.org}.}%
\thanks{$^{4}$Lihua Xie is with the NTU-VinUni Joint Research Laboratory for Robotics and Embodied AI, School of Electrical and Electronic Engineering, Nanyang Technological University, Singapore 639798, and also with VinUniversity, Hanoi 100000, Vietnam, {\tt lihua.x@vinuni.edu.vn}.}}

\newcommand{\anoncode}{The source code of AirLeg is available in an anonymized repository at \url{https://anonymous.4open.science/r/AirLeg-XXXX}.}

\markboth{}%
{}

\makeatletter
\def\anonauthor{Anonymous Author(s)\thanks{\anoncode}}
\def\noauthor{\@latex@warning@no@line{No \noexpand\author given}}
\newcommand{\namedonly}[1]{\ifx\@author\anonauthor\else\ifx\@author\noauthor\else#1\fi\fi}
\makeatother
\IEEEaftertitletext{%
  \vspace{-0.2\baselineskip}\namedonly{\vspace{-\baselineskip}}%
  \centering
  \setlength{\abovecaptionskip}{-0.01cm}
  \includegraphics[width=\textwidth]{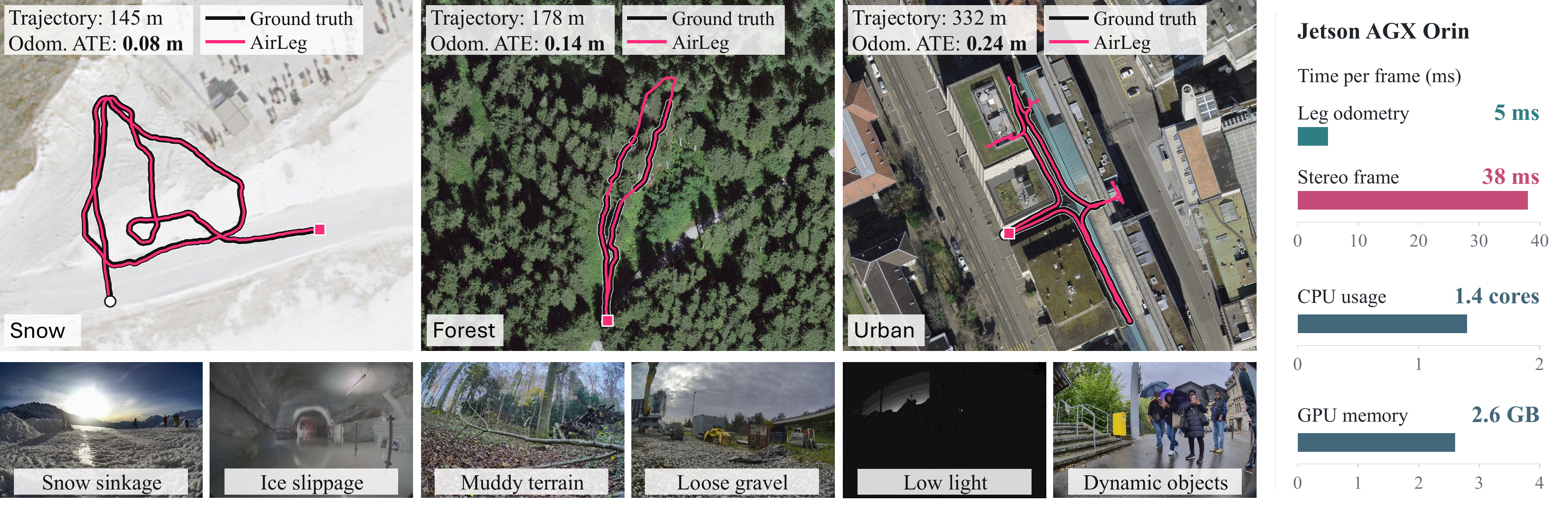}%
  \captionof{figure}{AirLeg on the GrandTour dataset.
  Top: trajectories estimated by AirLeg without loop closure and the ground truth on three sequences in snow, forest, and urban scenes.
  Bottom: challenging conditions in the test sequences, including snow sinkage, ice slippage, muddy terrain, loose gravel, low light, and dynamic objects.
  Right: runtime and resource usage on an NVIDIA Jetson AGX Orin.
  With its leg odometry adapted to the real robot without ground truth, AirLeg achieves accurate and robust localization in real time on an embedded computer.}%
  \label{fig:fig1}%
  \vspace{0.4\baselineskip}}
\maketitle

\begin{abstract}
Visual-inertial-leg (VIL) state estimation offers cost-effective localization for legged robots, adding only a camera to the onboard IMU and joint encoders.
However, learning-based leg odometry is limited by its supervision: networks trained in simulation suffer from the sim-to-real gap, while real-world labels require costly sensors or inherit the errors of visual estimates.
To address this, we present AirLeg, a self-supervised VIL SLAM system in which vision and leg proprioception improve each other from training to deployment.
We pre-train the leg odometry network in simulation and adapt it to the robot without ground truth through imperative learning, where multi-sensor bundle adjustment uses the leg odometry to refine the visual map, which in turn supervises the network.
During deployment, the fused estimates adapt the network online. To improve both localization and this supervision, a dual-window estimator fuses vision, IMU, and leg odometry, relates loaded feet to the visually reconstructed terrain through a deformation-aware constraint, and weights the constraints by variance component estimation.
AirLeg completes all 17 GrandTour sequences with the lowest average error (0.19\,m) among 19 methods.
On a low-cost quadruped, it achieves the lowest drift (0.6\%) with only 0.5\,h of unlabeled data, in real time on an embedded computer.
\end{abstract}


\IEEEpeerreviewmaketitle

\section{Introduction}

\IEEEPARstart{L}{egged} robots can traverse stairs, rubble, and uneven terrain, making them attractive for industrial inspection, search and rescue, and subterranean exploration \cite{hutter2016anymal}.
Translating this mobility into autonomous operation, however, requires accurate and robust localization to support mapping, planning, and motion control.
Among the possible sensor configurations, visual-inertial-leg (VIL) state estimation offers a promising and cost-effective approach to meeting this need by combining complementary sources of motion information.
Cameras provide rich geometric constraints and semantic information for drift correction.
Meanwhile, an inertial measurement unit (IMU) can bridge visual outages over short periods when vision degrades in dynamic scenes, texture-poor regions, or under challenging illumination \cite{cadena2016past, zuniga2020vi, xu2025airslam}.
Over longer periods in such conditions, where visual-inertial odometry (VIO) drifts rapidly \cite{forster2016manifold, qin2018vins}, joint encoder and contact measurements still provide motion information that is independent of visual appearance and does not require double integration \cite{bloesch2013state, hartley2020contact}.
Beyond this complementarity, VIL estimation also offers a practical advantage: legged robots are typically equipped with an IMU and joint encoders, and a camera is already needed in most robotic tasks for visual semantic information, so VIL estimation incurs very little additional sensor cost, making it an economical alternative to LiDAR-centric systems \cite{wisth2023vilens, ou2024legkilo}.

Despite this potential, achieving reliable VIL estimation in real-world environments remains challenging. When traversing difficult terrain, legged robots frequently experience foot impacts and rapid body motions that can degrade visual and inertial measurements \cite{wisth2023vilens}. Under such conditions, leg odometry becomes particularly valuable as a complementary source of motion information. However, its reliability is also affected by the complex robot-ground interactions encountered during terrain traversal. Conventional kinematics-based methods assume stationary stance feet and depend on an accurate kinematic model, making their estimates vulnerable to foot slip, ground deformation, and model errors \cite{bloesch2013state, hartley2020contact, camurri2020pronto, kim2022step, yang2022online}.

Learning-based methods relax these assumptions by training neural networks to regress the body velocity from proprioceptive measurements \cite{buchanan2022learning, youm2024legged, wasserman2025legolas}.
However, this approach shifts the difficulty to supervision, since the accuracy of the network depends on the data and labels it learns from.
Simulation provides abundant and diverse labeled data at low cost, but since actuator dynamics, sensor noise, and foot-ground contact cannot be simulated exactly, networks trained only in simulation degrade in accuracy on the physical robot \cite{youm2024legged, wasserman2025legolas}.
Training on real-world data \cite{buchanan2022learning, okawara2025tightly, kulkarni2026online} avoids this gap, but such data are laborious to collect across diverse terrains and gaits.
Moreover, existing methods face two challenges in supervision.
The first is \textbf{supervision cost}: the labels come from external equipment such as motion capture~\cite{buchanan2022learning} or from an additional LiDAR-based estimator~\cite{okawara2025tightly, kong2026trace}, even when the network is pre-trained in simulation and only fine-tuned on the robot~\cite{kong2026trace}.
The second is \textbf{estimator dependence}: an onboard visual-inertial estimator can replace these labels at no extra cost~\cite{kulkarni2026online}, but the network then learns only from the visual estimates and tends to reproduce them.
As a result, the errors of the visual estimates can propagate to the network, and the leg odometry cannot in turn improve its supervision, which reduces the benefit of fusing the two modalities when vision degrades.

These challenges lead to a key question: with neither additional sensors nor external ground-truth equipment, how can the leg odometry be adapted to the physical robot without being bound by the errors of the visual estimator, so that it complements vision rather than follows it?
Our key insight is that the leg odometry should not only learn from vision but also help to correct it, which leads to two principles. 
First, to keep the leg odometry complementary to vision, most of the network parameters should learn a motion prior of legged locomotion without relying on visual signals, rather than all being trained by visual supervision.
Second, to limit the errors inherited from vision, the supervision should be refined with multi-modal information rather than derived from vision alone.

Motivated by this, we present AirLeg, a self-supervised VIL SLAM system in which the leg odometry and the visual estimation improve each other from training to deployment.
To keep it complementary to vision, its leg odometry network is trained in three stages, with fewer trainable parameters in each stage.
It is first pre-trained in simulation, where all parameters are trained across diverse terrains, gaits, and robot parameters to learn the motion prior.
It is then adapted offline to the physical robot with a small amount of unlabeled real-world data, where only a small part of the parameters is fine-tuned.
To improve the visual supervision in the second stage, we formulate this offline adaptation as a bilevel optimization, building on imperative learning (IL)~\cite{wang2024imperative}.
At the lower level, a multi-sensor bundle adjustment (BA) uses the leg odometry to refine the visual map, so that the errors in the supervision are reduced.
At the upper level, the refined map in turn supervises the network through the reprojection errors of its points in the images rather than through pose regression, as these errors stay closer to the raw image measurements.
Unlike existing IL systems~\cite{fu2024islam, zhan2024imatching}, the two levels then have different objectives, so the gradient through the BA solution does not vanish, and we compute it with a differentiable leg odometry integration and an efficient sparse adjoint solve.
Finally, the network keeps learning online during deployment, where only a few low-rank parameters are added to fit the deployment environment.

During deployment, the legs are coupled with vision through both motion and geometry, so that the network is likewise supervised by multi-modal estimates rather than by vision alone.
To fuse the motion constraints over a long horizon at a low cost, we design a dual-window estimator: a short window optimizes the poses and map points of a leg-aided stereo visual odometry (VO), while a long window jointly optimizes a few high-quality landmarks with the proprioceptive motion constraints, with the remaining map points marginalized into relative-pose constraints.
To exploit the geometry between the feet and the terrain, we develop a deformation-aware foot-terrain constraint, in which the terrain is reconstructed from the stereo images and the leg kinematics places each loaded foot on it, while a slowly varying contact offset absorbs foot sinkage and systematic errors, which reduces the vertical drift.
The weights of the leg odometry and foot-terrain constraints are adjusted online by variance component estimation, and the fused estimates supervise the online adaptation of the network.
The main contributions of this work are as follows:
\begin{itemize}
  \vspace{-0.2em}
  \item We propose a self-supervised sim-to-real adaptation framework for learning-based leg odometry, which requires neither real-world ground truth nor additional sensors.
  The network is pre-trained in simulation and adapted to the physical robot, first offline and then online, with only a small amount of unlabeled real-world data.
  We formulate the offline adaptation as a bilevel optimization, in which a multi-sensor BA uses the leg odometry to refine the visual map that supervises the network.
  As its two levels have different objectives, we compute the direct and indirect gradients with a differentiable leg odometry integration and a sparse adjoint solve without unrolling BA.

  \item We develop a dual-window VIL estimator that couples the legs with vision through both motion and geometry.
  It fuses the leg odometry with vision and the IMU over a long horizon at a low cost, with most map points marginalized into relative-pose constraints, and ties each loaded foot to the visually reconstructed terrain through a deformation-aware constraint to limit the vertical drift.
  The weights of these constraints are adapted online by variance component estimation, and the fused estimates in turn supervise the online adaptation of the network.

  \item We extensively evaluate AirLeg on the large-scale GrandTour dataset against 18 open-source methods of four types, using only the built-in IMU of the robot, and in real-world tests on a low-cost Unitree Go2 quadruped.
 AirLeg achieves an average trajectory error of 0.19\,m on GrandTour and is the only method that completes all 17 test sequences with an error below 0.4\,m (Fig.~\ref{fig:fig1}), and on the Go2 it achieves an average drift of about 0.6\% of the traveled distance with only 0.5\,h of unlabeled real-world training data, while running in real time on an embedded computer.
  The results further show that the sim-to-real adaptation is data-efficient and transfers well across scene categories.
 We release the code and dataset.
\end{itemize}

\section{Related Work}\label{sec:related-work}

\subsection{Visual Localization}\label{sec:related-visual}
Visual simultaneous localization and mapping (SLAM) has been extensively studied over the past two decades \cite{cadena2016past}.
Classical systems estimate camera motion from handcrafted features or raw image intensities, using either filtering \cite{mourikis2007multi, geneva2020openvins} or nonlinear optimization \cite{engel2017direct, campos2021orb}, and are often tightly coupled with an IMU \cite{qin2018vins}.
They are accurate and efficient in well-textured and well-lit scenes, but their handcrafted front ends are fragile under low texture, motion blur, and illumination changes \cite{zuniga2020vi}, which are common during legged locomotion.
Learning-based methods address this fragility by embedding differentiable geometric optimization in recurrent architectures \cite{teed2021droid} or exploiting learned 3D reconstruction priors \cite{murai2025mast3r}.
They achieve state-of-the-art robustness on many benchmarks, but usually demand substantial GPU resources and are less straightforward to couple with additional sensing modalities.
Hybrid systems combine the strengths of both paradigms by using learned feature detection and matching in the front end and classical geometric optimization in the back end \cite{detone2018superpoint, lindenberger2023lightglue, xu2023airvo, xu2025airslam}.
A recent study \cite{cioffi2026why} further reveals that the advantage of deep visual SLAM mainly stems from learned 2D data association and uncertainty rather than from the recurrent architecture itself, lending support to this hybrid design.
Our system follows this paradigm and builds its visual component upon AirSLAM \cite{xu2025airslam}.
Nevertheless, any purely visual system inevitably degrades when informative visual observations are absent, motivating the incorporation of complementary modalities, such as the proprioceptive sensing of legged robots.

\subsection{Leg Odometry and Leg-Aided State Estimation}\label{sec:related-leg}
Conventional leg odometry computes the body velocity from joint encoders and forward kinematics under the assumption that stance feet remain stationary, and fuses it with IMU measurements via extended Kalman filters \cite{bloesch2013state, camurri2020pronto}, invariant extended Kalman filters \cite{hartley2020contact}, or factor graphs with preintegrated foot velocity factors \cite{kim2022step}.
Building on these formulations, multi-sensor systems further incorporate vision \cite{yang2023cerberus} or LiDAR \cite{wisth2023vilens, ou2024legkilo} to reduce long-term drift.
However, their accuracy hinges on the no-slip assumption, reliable contact detection \cite{camurri2017probabilistic}, and an accurate kinematic model.
Although slip rejection and online kinematic calibration \cite{yang2022online, yang2023cerberus} mitigate some of these errors, they rely on hand-designed rules and remain vulnerable on deformable terrain, where the feet persistently sink and slide during the stance phases that are assumed static.

Learning-based methods relax these modeling assumptions.
Early works learn individual components, such as contact events \cite{lin2021legged} or inertial displacements \cite{buchanan2022learning}, and integrate them into model-based estimators, while more recent works regress the body velocity or odometry directly from proprioceptive measurements \cite{youm2024legged, wasserman2025legolas}.
To cope with changing terrains and payloads, some systems further train the network online, supervised by LiDAR-inertial \cite{okawara2025tightly} or other exteroceptive estimates \cite{kulkarni2026online}.
Despite their promising results, these methods face a dilemma in obtaining supervision.
Networks trained purely in simulation \cite{youm2024legged, wasserman2025legolas} are inexpensive to train but suffer from the sim-to-real gap, whereas networks trained on real-world data require either costly ground-truth equipment or exteroceptive supervision whose errors are inherited by the network.
Moreover, when supervised online by exteroceptive estimates, leg odometry tends to converge toward its teacher and thus inherits its errors, particularly during exteroceptive degradation.
Although degraded estimates can be excluded from the supervision~\cite{kulkarni2026online}, visual degradation has various causes and is a matter of degree, so it is hard to separate with fixed thresholds.
Finally, both conventional and learning-based systems typically fuse leg information only as relative motion or velocity constraints, leaving the geometric relationship between the feet and the observed terrain unexploited.

\subsection{Imperative Learning}\label{sec:related-il}
Imperative learning (IL) \cite{wang2024imperative} is a self-supervised neuro-symbolic framework formulated as a bilevel optimization problem.
In the upper level, a neural network produces predictions, while in the lower level, a physics- or geometry-based reasoning module, such as an optimizer or a planner, refines these predictions under explicit constraints.
The lower-level solution in turn provides supervision for the network, so that both levels are improved without external labels.
IL has been successfully applied to path planning \cite{yang2023iplanner, chen2025ia}, feature matching \cite{zhan2024imatching}, and visual SLAM \cite{fu2024islam}, where the network learns exteroceptive perception or planning modules from its own geometric reasoning.
Different from these works, we employ IL to adapt a simulation-pretrained proprioceptive model to the real robot.
In our formulation, the lower-level multi-sensor bundle adjustment refines the visual geometry using the leg odometry predictions, and the refined geometry supervises the leg odometry network through visual reprojection and epipolar errors.
This cross-modal bilevel formulation allows the visual and proprioceptive modalities to improve each other, which, to the best of our knowledge, has not been explored in prior IL methods or learning-based leg odometry.

\section{Simulation Pre-training}\label{sec:sim}
In this section, we introduce the first stage of our training framework, which pre-trains the leg odometry network entirely in simulation.
Unlike learning-based leg odometry methods that rely solely on real-world training data, we use simulation to reduce the cost of learning motion patterns across diverse conditions.
Training for broad generalization requires data spanning different terrains, gaits, speeds, and payloads, but collecting such data on physical robots with accurate ground truth is laborious and expensive.
Simulation makes it possible to generate these labeled data at scale and at low cost, while also covering conditions that are difficult or unsafe to reproduce in real-world experiments.
Through this pre-training, the network learns transferable relationships between proprioceptive measurements and body motion, providing a foundation for adaptation to the physical robot using a small amount of real-world data in the next stage.

\subsection{Network Design}\label{sec:sim-network}
\begin{figure}[t]
  \centering
  \setlength{\abovecaptionskip}{-0.5cm}
  \includegraphics[width=\columnwidth]{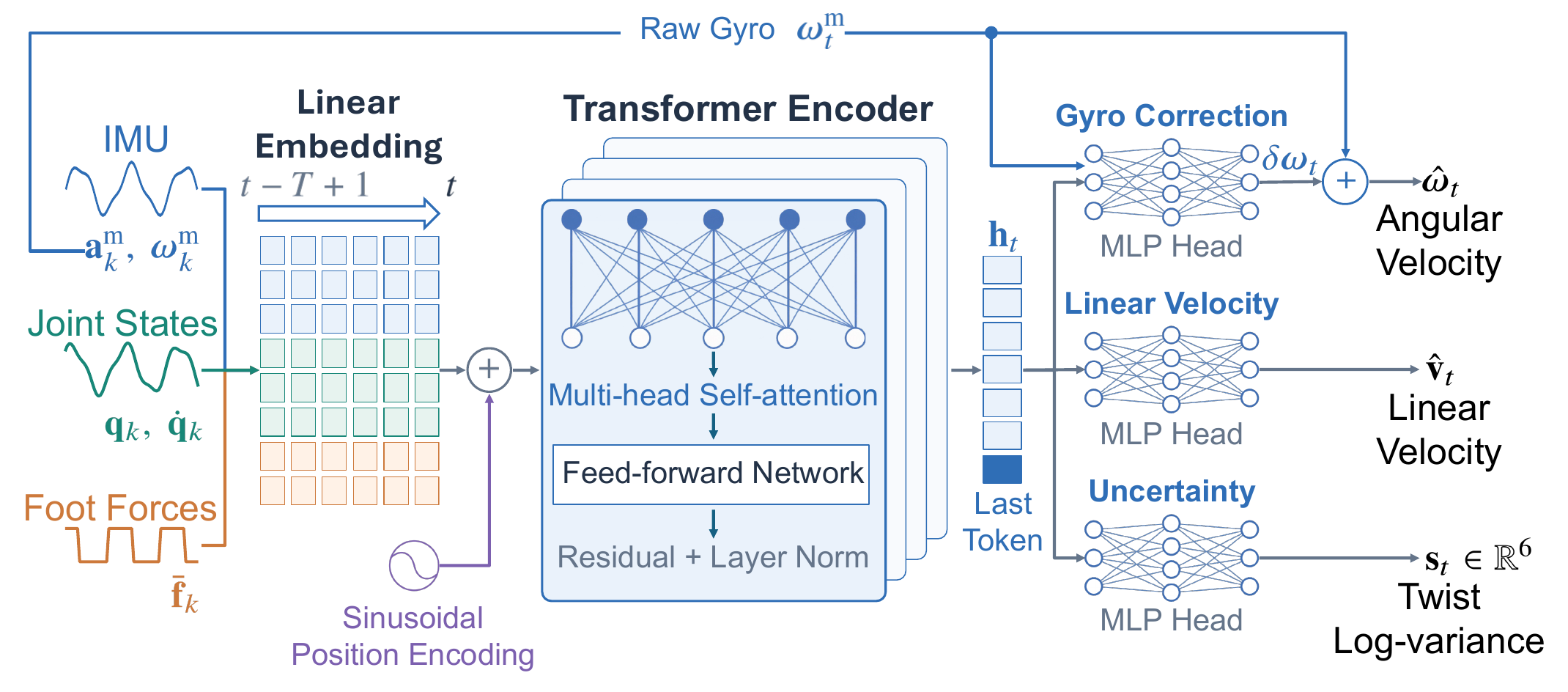}
  \caption{Architecture of the leg odometry network.
  A temporal window of IMU measurements, joint positions and velocities, and normalized foot forces is embedded into tokens and processed by a Transformer encoder.
  The last-token feature is passed to three MLP heads that predict the body-frame linear velocity, body-twist uncertainty, and an angular-velocity correction.
  The gyro-correction head additionally receives the raw gyroscope measurement.}
  \vspace{-0.3em}
  \label{fig:network}
\end{figure}

Our network estimates the robot's body velocity from a short history of proprioceptive measurements.
We use velocity rather than pose as the learning target because it captures local motion without depending on the robot's global position or heading, facilitating generalization across trajectories.
Given a causal window of $T = 50$ measurement samples, the network predicts the body-frame twist $\bm{\xi}_t = [\mathbf{v}_t^{\top}, \bm{\omega}_t^{\top}]^{\top}$ at time $t$ and its uncertainty, as illustrated in Fig.~\ref{fig:network}.
Each input sample contains IMU measurements, joint positions and velocities, and foot contact forces, providing information about both body motion and foot-ground interactions.
For robots without reliable foot force sensors, we estimate the contact forces from joint torques.
Specifically, given the joint torques $\bm{\tau}_i$ and foot Jacobian $\mathbf{J}_i$ of leg $i$, we compute the force using damped least squares:
\begin{equation}
  \mathbf{F}_i = \left(\mathbf{J}_i\mathbf{J}_i^{\top} + \lambda^2\mathbf{I}\right)^{-1}\mathbf{J}_i\bm{\tau}_i,
  \label{eq:foot_force}
\end{equation}
where $\lambda$ is a small damping factor.
The magnitude of its $z$-component, $f_i = |F_{i,z}|$, is used as the contact feature.

However, the force magnitudes depend on the robot weight and payload, and also differ between simulated and real actuators.
We therefore normalize them by an estimate of the total supported load:
\begin{equation}
  \bar{f}_{k,i} = \frac{f_{k,i}}{\max\left(\bar{P}_k, P_{\min}\right)},
  \label{eq:force_norm}
\end{equation}
where $\bar{P}_k$ is the time-weighted mean of the total force $\sum_i f_{k,i}$ over a recent time window, and $P_{\min}$ prevents division by a near-zero load estimate.
The normalized features thus capture how the load is distributed across the legs rather than its overall scale, which facilitates the transfer of the network across different payloads and from simulation to reality.

To capture the temporal structure of joint motions and contacts, we linearly embed each input vector into a token and add a sinusoidal positional encoding.
The resulting sequence is processed by a Transformer encoder with four layers and four attention heads.
The last-token feature, $\mathbf{h}_t$, represents the measurement history and is passed to three multilayer perceptron (MLP) heads.
The velocity head predicts the body-frame linear velocity $\hat{\mathbf{v}}_t$, while the uncertainty head predicts six log-variances for the body twist.
Since the gyroscope already provides a direct measurement of angular velocity, the gyro-correction head predicts a residual correction rather than the full angular velocity.
It takes $\mathbf{h}_t$ and the raw gyroscope measurement $\bm{\omega}_t^{\mathrm{m}}$ as input and outputs $\delta\bm{\omega}_t$, yielding
$\hat{\bm{\omega}}_t = \bm{\omega}_t^{\mathrm{m}} + \delta\bm{\omega}_t$.
The complete network has about 0.85\,M parameters.

\subsection{Simulation Data and Training}\label{sec:sim-data}
We generate the training data in Isaac Lab~\cite{mittal2023orbit} with two types of learned locomotion policies: gait-conditioned policies~\cite{margolis2023walk} that trot, pace, bound, and pronk under random velocity commands, and terrain-traversal policies~\cite{cheng2024extreme} that negotiate slopes, stairs, gaps, hurdles, and parkour courses.
Both are run on rigid ground with varying friction, gravel, and soft compliant ground, while the robot and actuator parameters, external disturbances, and sensor noise and biases are randomized, as summarized in Table~\ref{tab:dr}.
For consistency with deployment, foot forces are estimated from the simulated joint torques and normalized by \eqref{eq:foot_force} and \eqref{eq:force_norm}.
In total, we collect about 80 hours of data per robot to train the network.

\begin{table}[t]
  \centering
  \caption{Simulation settings.}
  \label{tab:dr}
  \renewcommand{\arraystretch}{1.1}
  \begin{tabular}{@{}l p{6.1cm}@{}}
    \toprule
    Item & Settings \\
    \midrule
    Terrains & Flat, slopes, stairs, gaps, hurdles, and mixed parkour courses, each with 10 difficulty levels \\
    Ground & Rigid ($\mu \in \{0.4, 0.6, 0.8, 1.25\}$), gravel (2--8\,cm), and soft ground (two stiffness levels) \\
    Gaits & Trot, pace, bound, and pronk \\
    Robot & Base mass $\times[0.85, 1.15]$, CoM offset $\pm 2$\,cm, limb mass $\times[0.9, 1.1]$ \\
    Actuators & Gains $\times[0.7, 1.3]$, strength $\times[0.85, 1.1]$, joint friction, armature, zero offsets, and delays \\
    Disturbances & Random external forces, torques, and pushes \\
    Sensor noise & Gyro.\ $0.03$\,rad/s (bias $0.01$), accel.\ $0.05$\,m/s$^2$ (bias $0.02$), joint position / velocity / torque $0.01$\,rad / $0.03$\,rad/s / $0.05$\,N\,m \\
    \bottomrule
  \end{tabular}
\end{table}
\begin{figure*}[t]
  \centering
  \setlength{\abovecaptionskip}{-0.2cm}
  \includegraphics[width= 0.9\linewidth]{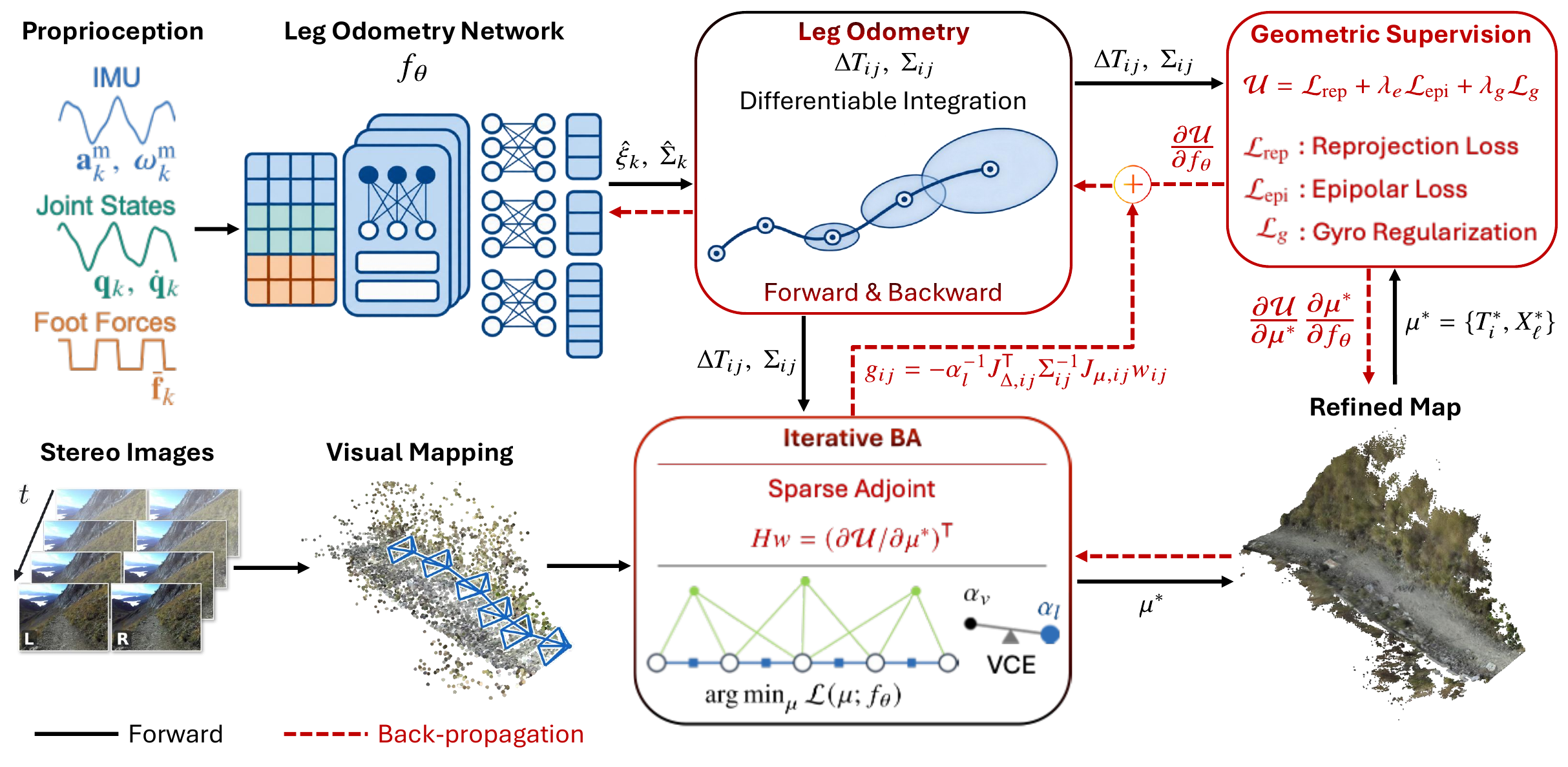}
  \caption{Overview of imperative sim-to-real adaptation.
  In the forward pass (solid arrows), the predictions of the leg odometry network $f_{\bm{\theta}}$ are integrated into relative poses with covariances and fused with the visual factors in lower-level BA, weighted by variance component estimation (VCE), and the refined map $\bm{\mu}^{*}$ supervises the network through the upper-level objective $\mathcal{U}$.
  In back-propagation (dashed arrows), the full gradient of $\mathcal{U}$ sums a direct term through the differentiable integration and an indirect term through the refined map, obtained as $\mathbf{g}_{ij}$ by a sparse adjoint solve without unrolling BA.}
  \label{fig:system}
  \vspace{-0.5em}
\end{figure*}
We jointly train the motion and uncertainty heads using a heteroscedastic Gaussian negative log-likelihood loss:
\begin{equation}
  \mathcal{L} = \frac{1}{2}\sum_{j=1}^{6}\left[\left(\hat{\xi}_{t,j} - \xi_{t,j}\right)^2 e^{-s_{t,j}} + s_{t,j}\right],
  \label{eq:loss_nll}
\end{equation}
where $\hat{\bm{\xi}}_t = [\hat{\mathbf{v}}_t^{\top}, \hat{\bm{\omega}}_t^{\top}]^{\top}$ is the predicted twist and $s_{t,j}$ is the predicted log-variance of its $j$-th component.
The first term weights the squared error by the predicted precision, and the second penalizes overly large variances.
The network thus learns accurate twists together with uncertainties that reflect its remaining errors, which for the angular components are those after the learned gyroscope correction.

\section{Sim-to-Real Adaptation}\label{sec:sim2real}

Despite domain randomization, the network trained purely in simulation still faces a sim-to-real gap on the physical robot: actuator dynamics, sensor noise and biases, kinematic calibration errors, and foot-ground contact on real terrains cannot be modeled exactly in simulation.
These discrepancies can introduce systematic velocity errors that accumulate into odometry drift and bias the leg constraints used in multi-sensor fusion.
We therefore adapt the simulation-pretrained network using real-world data to close this gap. 
To obtain high-quality supervision without additional ground-truth equipment or higher-precision sensors, we jointly refine the network and its supervisory signal through alternating optimization.
We first formulate the adaptation as an imperative learning problem with two coupled optimization levels in Section~\ref{sec:s2-formulation}.
We then introduce the differentiable integration that links these levels in Section~\ref{sec:s2-integration} and detail their respective optimization procedures in Sections~\ref{sec:s2-lower} and~\ref{sec:s2-upper}.

\subsection{Problem Formulation}\label{sec:s2-formulation}

Our goal is to adapt the leg odometry network on the physical robot using only onboard stereo images and proprioceptive measurements.
A straightforward approach is to obtain camera poses or velocities from SLAM or structure from motion (SFM) and use these estimates as fixed pseudo labels.
However, their accuracy depends on the quality of the visual observations.
Texture-poor regions, motion blur induced by foot impacts, and dynamic objects can introduce noise and bias into the visual estimates, which may then be transferred to the network during training.
To reduce this dependence on fixed, potentially inaccurate labels, we formulate the sim-to-real adaptation as an imperative learning problem~\cite{wang2024imperative}, in which the network and the visual estimates that supervise it are refined together through alternating optimization.

Specifically, we first run AirSLAM \cite{xu2025airslam} on each recorded sequence to obtain the initial feature associations and map points.
Let $\bm{\mu} = \{\mathbf{T}_i, \mathbf{X}_l\}$ denote the keyframe poses $\mathbf{T}_i \in SE(3)$ and map points $\mathbf{X}_l \in \mathbb{R}^3$, and let $f_{\bm{\theta}}$ denote the leg odometry network with parameters $\bm{\theta}$.
Here, $\bm{\theta}$ comprises only the prediction heads and a small part of the backbone, while the other parameters are frozen to preserve the features learned in simulation.
We formulate the adaptation as
\begin{subequations}\label{eq:bilevel}
  \begin{align}
    \min_{\bm{\theta}}\quad & \mathcal{U}\big(f_{\bm{\theta}};\bm{\mu}^{*}\big), \label{eq:bilevel_upper}\\
    \text{s.t.}\quad & \bm{\mu}^{*}=\arg\min_{\bm{\mu}}\ \mathcal{L}\big(\bm{\mu};f_{\bm{\theta}}\big), \label{eq:bilevel_lower}
  \end{align}
\end{subequations}
where the lower-level $\mathcal{L}$ is a multi-sensor bundle adjustment (BA) objective that combines visual observations with the predicted leg odometry by $f_{\bm{\theta}}$, and the upper-level $\mathcal{U}$ measures how well the network-predicted motion agrees with the refined visual geometry.
Intuitively, at the lower level, BA estimates the most likely poses and map points given the current network, whereas at the upper level, the network is optimized to be geometrically consistent with the refined map.
As illustrated in Fig.~\ref{fig:system}, this interaction allows improved leg odometry to support map refinement, while the refined map provides updated supervision for subsequent network training.

\subsection{Differentiable Leg Odometry Integration}\label{sec:s2-integration}

Connecting the two optimization levels requires converting network predictions into keyframe-level motion constraints while preserving a gradient path for training.
In the forward pass, predicted velocities and uncertainties are integrated into relative poses and covariances for BA.
In the backward pass, gradients of the upper-level loss propagate through these relative poses to the network.
We implement this operation as a differentiable integration layer on the Lie group~\cite{wang2023pypose} and use the same linearization for covariance propagation and gradient computation.
Specifically, let $t_i$ and $t_j$ denote the timestamps of consecutive keyframes, and let $k$ index the network predictions between them.
Each prediction covers an interval $\Delta t_k$, truncated at the keyframe boundaries where necessary.
Starting from $\Delta\mathbf{R}_{ii}=\mathbf{I}$ and $\Delta\mathbf{p}_{ii}=\mathbf{0}$, we integrate the predicted twists as
\begin{equation}
  \begin{aligned}
    \Delta\mathbf{R}_{i,k+1} &= \Delta\mathbf{R}_{ik}\,\mathrm{Exp}\big(\hat{\bm{\omega}}_k\Delta t_k\big), \\
    \Delta\mathbf{p}_{i,k+1} &= \Delta\mathbf{p}_{ik}+\Delta\mathbf{R}_{ik}\,\hat{\mathbf{v}}_k\Delta t_k,
  \end{aligned}
  \label{eq:integration}
\end{equation}
where the angular velocity $\hat{\bm{\omega}}_k = \bm{\omega}^{\mathrm{m}}_k - \mathbf{b}_g + \delta\bm{\omega}_k$ is corrected by a constant gyroscope bias $\mathbf{b}_g$, which is estimated beforehand from static intervals and long-baseline visual rotations, and by the learned correction $\delta\bm{\omega}_k$.
Using the camera-to-body extrinsic transformation $\mathbf{T}_{BC}$, we convert the integrated body-frame motion $\Delta\mathbf{T}^{B}_{ij}$ to the camera frame:
$\Delta\mathbf{T}_{ij}=\mathbf{T}_{BC}^{-1}\Delta\mathbf{T}^{B}_{ij}\mathbf{T}_{BC}$.

We linearize \eqref{eq:integration} using a right-multiplicative rotation perturbation,
$\Delta\mathbf{R}_{ik}\leftarrow\Delta\mathbf{R}_{ik}\mathrm{Exp}(\delta\bm{\phi}_k)$,
and an additive position perturbation.
The resulting error state $\bm{\epsilon}_k=[\delta\bm{\phi}_k^{\top},\delta\mathbf{p}_k^{\top}]^{\top}$ evolves according to
\begin{equation}
  \begin{aligned}
    \bm{\epsilon}_{k+1} &= \mathbf{A}_k\,\bm{\epsilon}_k+
    \mathbf{B}_k
    \begin{bmatrix}
      \delta\hat{\mathbf{v}}_k\\
      \delta\hat{\bm{\omega}}_k
    \end{bmatrix},\\
    \mathbf{A}_k &=
    \begin{bmatrix}
      \mathrm{Exp}(-\hat{\bm{\omega}}_k\Delta t_k) & \mathbf{0}\\
      -\Delta\mathbf{R}_{ik}[\hat{\mathbf{v}}_k]_{\times}\Delta t_k & \mathbf{I}
    \end{bmatrix},\\
    \mathbf{B}_k &=
    \begin{bmatrix}
      \mathbf{0} & \mathbf{J}_r(\hat{\bm{\omega}}_k\Delta t_k)\Delta t_k\\
      \Delta\mathbf{R}_{ik}\Delta t_k & \mathbf{0}
    \end{bmatrix},
  \end{aligned}
  \label{eq:error_state}
\end{equation}
where $[\cdot]_{\times}$ denotes the skew-symmetric matrix and $\mathbf{J}_r$ is the right Jacobian of $SO(3)$.
In the forward pass, this linearization propagates the predicted velocity uncertainties into the relative-motion covariance:
\begin{equation}
  \mathbf{P}_{k+1}=
  \mathbf{A}_k\mathbf{P}_k\mathbf{A}_k^{\top}+
  \mathbf{B}_k\,\mathrm{diag}\big(\hat{\bm{\sigma}}^2_k\big)\,\mathbf{B}_k^{\top}.
  \label{eq:cov_prop}
\end{equation}
We then transform the covariance to the camera frame using the corresponding adjoint transformation $\mathbf{T}_{BC}$ and add a small noise floor to obtain the leg-odometry covariance $\bm{\Sigma}_{ij}$.
In the backward pass, the same linearization gives the sensitivity of the integrated motion to each prediction:
\begin{equation}
  \frac{\partial\bm{\epsilon}_{j}}
  {\partial[\hat{\mathbf{v}}_k;\hat{\bm{\omega}}_k]}
  =\mathbf{A}_{j-1}\cdots\mathbf{A}_{k+1}\mathbf{B}_k.
  \label{eq:int_jacobian}
\end{equation}
This Jacobian propagates the upper-level gradient to all predictions within the interval, allowing the linear velocity and gyroscope correction to be updated from geometric supervision.

\subsection{Lower-Level Optimization}\label{sec:s2-lower}

\subsubsection{Bundle Adjustment}

Given the relative motions $\Delta\mathbf{T}_{ij}$ and covariances $\bm{\Sigma}_{ij}$ from Section~\ref{sec:s2-integration}, the lower level jointly refines the keyframe poses and map points by minimizing
\begin{equation}
  \mathcal{L}(\bm{\mu};f_{\bm{\theta}})
  =\sum\nolimits_{i,l}\tfrac{1}{\alpha_{\mathrm{v}}}
  \|\pi(\mathbf{T}_i^{-1}\mathbf{X}_l)-\mathbf{x}_{il}\|^2
  +\sum\nolimits_{i,j}\tfrac{1}{\alpha_{\mathrm{l}}}
  \|\mathbf{r}_{ij}\|^2_{\bm{\Sigma}_{ij}^{-1}},
  \label{eq:lower}
\end{equation}
where $\pi(\cdot)$ is the image projection function, $\mathbf{x}_{il}$ is the observation of map point $l$ in keyframe $i$, and
$\mathbf{r}_{ij}=\mathrm{Log}\big(\Delta\mathbf{T}_{ij}^{-1}\mathbf{T}_i^{-1}\mathbf{T}_j\big)$
is the leg-odometry residual.
We initialize the optimization with the AirSLAM map and reject visual observations with large reprojection errors as outliers during optimization.

\subsubsection{Variance Component Estimation}

Since the appropriate variance scales $\alpha_{\mathrm{v}}$ and $\alpha_{\mathrm{l}}$ vary with terrain, robot motion, and visual conditions, we calibrate them from the BA residuals by variance component estimation (VCE)~\cite{teunissen2008least}.
Specifically, for each observation group $g\in\{\mathrm{v},\mathrm{l}\}$, we model its covariance as $\alpha_g\bm{\Sigma}^{0}_g$, where $\bm{\Sigma}^{0}_g$ is the prior covariance and $\alpha_g$ is an unknown scale.
The visual prior $\bm{\Sigma}^{0}_{\mathrm{v}}$ describes keypoint pixel noise, while the leg prior $\bm{\Sigma}^{0}_{\mathrm{l}}$ is assembled from the propagated covariances $\bm{\Sigma}_{ij}$.
Given the current scales, let $\mathbf{e}_g$ and $\mathbf{J}_g$ denote the residual vector and Jacobian at the BA solution, whitened using $\alpha_g\bm{\Sigma}^{0}_g$, and let $m_g$ be the number of scalar residuals in group $g$.
The corresponding Gauss--Newton Hessian is $\mathbf{H}=\sum_g\mathbf{H}_g$, where $\mathbf{H}_g=\mathbf{J}_g^{\top}\mathbf{J}_g$.
Since fitting the states absorbs part of the observation noise, the expected residual energy of a group is determined not by the residual count $m_g$ but by its redundancy, \ie the remaining degrees of freedom:
\begin{equation}
  \nu_g=m_g-\mathrm{tr}\big(\mathbf{H}^{-1}\mathbf{H}_g\big),
  \label{eq:redundancy}
\end{equation}
where the trace term accounts for the degrees of freedom of group $g$ that are used in fitting the states.

To estimate the scale correction, let $\beta_g$ denote the ratio of the true variance scale to its current value.
Under the linearized model, the expected residual energy is a linear function of $\bm{\beta}=[\beta_{\mathrm{v}},\beta_{\mathrm{l}}]^{\top}$:
\begin{equation}
  \begin{gathered}
    \mathbb{E}\big[\mathbf{e}_g^{\top}\mathbf{e}_g\big]=\sum_h S_{gh}\,\beta_h,\\
    S_{gh}=\delta_{gh}\big(m_g\!-\!2\,\mathrm{tr}(\mathbf{H}^{-1}\mathbf{H}_g)\big)\!+\!\mathrm{tr}\big(\mathbf{H}^{-1}\mathbf{H}_g\mathbf{H}^{-1}\mathbf{H}_h\big),
  \end{gathered}
  \label{eq:vce}
\end{equation}
where $\delta_{gh}$ is the Kronecker delta.
Replacing the expected energies with the observed values $q_g=\mathbf{e}_g^{\top}\mathbf{e}_g$ gives
\begin{equation}
  \hat{\bm{\beta}}=\mathbf{S}^{-1}\mathbf{q},
  \qquad \mathbf{q}=[q_{\mathrm{v}},q_{\mathrm{l}}]^{\top}.
  \label{eq:vce_solve}
\end{equation}
Intuitively, $\hat{\beta}_g\approx q_g/\nu_g$, \ie a group is down-weighted when its residual energy exceeds that expected from its redundancy, while \eqref{eq:vce_solve} also accounts for the coupling between groups.

Since the scales in turn affect the BA solution and its Hessian, we alternate between scale estimation and BA.
At each iteration, we apply a damped update in the logarithmic domain:
\begin{equation}
  \alpha_g\leftarrow\alpha_g\exp\Big(
  \mathrm{clip}\big(\eta\log\hat{\beta}_g,-\log\gamma,\log\gamma\big)
  \Big),
  \label{eq:vce_update}
\end{equation}
where $\eta\in(0,1]$ controls the damping and $\gamma$ bounds the multiplicative change in each step.
The iterations stop when all $|\log\hat{\beta}_g|$ fall below a small threshold.
An update is rejected and the current scales are retained if $\mathbf{S}$ is ill-conditioned or any estimated ratio $\hat{\beta}_g$ is non-positive.
To capture changes in sensor reliability over time, we estimate the scales in local keyframe windows and interpolate between window centers in the logarithmic domain.
Windows that fail to converge use unit scales.
A final BA pass with the resulting time-varying weights produces the refined map $\bm{\mu}^{*}$ for upper-level training.

\subsection{Upper-Level Optimization}\label{sec:s2-upper}

\subsubsection{Objective}

At the upper level, we use the map refined by lower-level BA to adapt the leg odometry network through geometric supervision.
Instead of directly regressing its predictions toward the refined poses, we train the network by reprojecting the refined map points into the observed images using its predicted motion.
This choice is motivated by two considerations.
First, reprojection errors naturally reflect how strongly the visual geometry constrains each motion direction.
In contrast, an unweighted pose-regression loss can impose equally strong supervision along poorly observed directions, such as motion along the viewing direction in distant scenes.
Second, map points are triangulated from multiple keyframes, and individual observations can be robustly down-weighted.
This keeps the loss tied to image evidence and reduces direct dependence on refined pose targets that are themselves partly determined by the network's own leg odometry predictions.

Specifically, for a keyframe pair $(i,j)$, let $\mathbf{X}^{i}_l=\mathbf{T}_i^{*-1}\mathbf{X}^{*}_l$ denote a refined map point expressed in the camera frame of keyframe $i$, with observation $\mathbf{x}_{jl}$ in keyframe $j$.
We define the reprojection loss as
\begin{equation}
  \mathcal{L}_{\mathrm{rep}}
  =\sum_{(i,j)}\sum_l
  \rho\Big(\pi\big(\Delta\mathbf{T}_{ij}^{-1}\,\mathbf{X}^{i}_l\big)-\mathbf{x}_{jl}\Big),
  \label{eq:loss_rep}
\end{equation}
where $\rho(\cdot)$ is the Huber loss.
We evaluate this loss in both directions for each pair.
For matched keypoints without a valid refined map point, we instead use an epipolar loss $\mathcal{L}_{\mathrm{epi}}$ based on the Sampson distance and the fundamental matrix induced by $\Delta\mathbf{T}_{ij}$.
The training pairs include both consecutive keyframes and nonconsecutive pairs separated by longer intervals within short keyframe windows.
For the latter, relative motions are composed from the network predictions, extending the supervision to longer temporal intervals.
The complete upper-level objective is
\begin{equation}
  \mathcal{U}
  =\mathcal{L}_{\mathrm{rep}}
  +\lambda_{\mathrm{e}}\mathcal{L}_{\mathrm{epi}}
  +\lambda_{\mathrm{g}}\mathcal{L}_{\mathrm{g}},
  \label{eq:upper}
\end{equation}
where $\mathcal{L}_{\mathrm{g}}$ regularizes the magnitude, temporal variation, and mean of $\delta\bm{\omega}$ to discourage the gyro head from compensating for errors unrelated to the gyroscope.

\subsubsection{Sparse Adjoint Differentiation}
The network predictions affect the upper-level objective in two ways: directly through the integrated motion, and indirectly through their influence on the map refined by lower-level BA. By the chain rule,
\begin{equation}
  \frac{\mathrm{d}\mathcal{U}}{\mathrm{d}\bm{\theta}}
  =\left(
    \frac{\partial\mathcal{U}}{\partial f_{\bm{\theta}}}
    +\frac{\partial\mathcal{U}}{\partial\bm{\mu}^{*}}
    \frac{\partial\bm{\mu}^{*}}{\partial f_{\bm{\theta}}}
  \right)
  \frac{\partial f_{\bm{\theta}}}{\partial\bm{\theta}}.
  \label{eq:chain}
\end{equation}
The gradient consists of a direct contribution that back-propagates through the differentiable integration via \eqref{eq:int_jacobian} and an indirect contribution arising from the dependence of the refined map on the network predictions.
Unlike in~\cite{fu2024islam,zhan2024imatching}, the stationarity condition of the lower-level problem does not eliminate the indirect term in our formulation.
This is because $\mathcal{U}$ evaluates geometric consistency using the network-predicted motion, whereas $\mathcal{L}$ uses the optimized poses.
Consequently, $\partial\mathcal{L}/\partial\bm{\mu}^{*}=0$ does not imply $\partial\mathcal{U}/\partial\bm{\mu}^{*}=0$, and the indirect contribution must also be considered.
To compute this indirect contribution without back-propagating through the BA iterations, we use implicit differentiation to determine how the optimized map changes with the network predictions:
\begin{equation}
  \frac{\partial\mathcal{U}}{\partial\bm{\mu}^{*}}
  \frac{\partial\bm{\mu}^{*}}{\partial f_{\bm{\theta}}}
  =-\mathbf{w}^{\top}
  \frac{\partial^2\mathcal{L}}{\partial\bm{\mu}\,\partial f_{\bm{\theta}}},
  \qquad
  \mathbf{H}\mathbf{w}
  =\Big(\frac{\partial\mathcal{U}}{\partial\bm{\mu}^{*}}\Big)^{\!\top},
  \label{eq:adjoint}
\end{equation}
where $\mathbf{H}$ is the lower-level Hessian, approximated in practice by the Gauss--Newton matrix evaluated at the current BA solution, and $\mathbf{w}$ is the adjoint vector given by the second equation.

To evaluate the resulting gradient, we exploit the structure of the leg odometry factors, through which the network predictions enter the lower-level optimization.
Each factor depends only on its predicted relative motion $\Delta\mathbf{T}_{ij}$ and the two connected poses.
Thus, after solving for $\mathbf{w}$, we compute the indirect gradient with respect to a right perturbation of each relative motion using the corresponding factor Jacobians and entries of $\mathbf{w}$:
\begin{equation}
  \mathbf{g}_{ij}
  =-\tfrac{1}{\alpha_{\mathrm{l}}}\,
  \mathbf{J}_{\Delta,ij}^{\top}
  \bm{\Sigma}_{ij}^{-1}
  \mathbf{J}_{\mu,ij}\,\mathbf{w}_{ij},
  \label{eq:implicit_g}
\end{equation}
where $\mathbf{J}_{\mu,ij}$ and $\mathbf{J}_{\Delta,ij}$ are the Jacobians of $\mathbf{r}_{ij}$ with respect to the two connected poses and a right perturbation of $\Delta\mathbf{T}_{ij}$, respectively, and $\mathbf{w}_{ij}$ contains the corresponding entries of the adjoint vector.
A single sparse linear solve therefore suffices to compute the indirect gradients for all leg odometry factors, with a cost comparable to one BA iteration.
During differentiation, we hold the factor weights, including the VCE scales, and the inlier set fixed at the values used in the lower-level optimization.

\subsubsection{Gradient Correction}

Since solving BA at every training iteration would be expensive, in practice we re-solve it only every few epochs.
Specifically, each round solves BA with the current network, computes the indirect gradients $\mathbf{g}_{ij}$ through \eqref{eq:implicit_g}, and then updates the network for several epochs with the refined map held fixed.
Within a round, however, the map and $\mathbf{g}_{ij}$ remain linearized at the relative motions $\Delta\mathbf{T}^{0}_{ij}$ used in BA, whereas the network predictions $\Delta\mathbf{T}_{ij}$ keep changing, and this inconsistency introduces errors in the gradients.
To address this, we approximate the response of the map to the predictions to first order around $\Delta\mathbf{T}^{0}_{ij}$, which yields the surrogate objective
\begin{equation}
  \tilde{\mathcal{U}}
  =\mathcal{U}
  +\sum\nolimits_{i,j}
  \big\langle\mathbf{g}_{ij},\,
  \mathrm{Log}\big(
    (\Delta\mathbf{T}^{0}_{ij})^{-1}\Delta\mathbf{T}_{ij}
  \big)\big\rangle,
  \label{eq:surrogate}
\end{equation}
where $\Delta\mathbf{T}^{0}_{ij}$ and $\mathbf{g}_{ij}$ are held constant within the round, while $\Delta\mathbf{T}_{ij}$ is computed from the current network predictions.
The logarithmic term measures the deviation of the predicted relative motion from the reference $\Delta\mathbf{T}^{0}_{ij}$; at $\Delta\mathbf{T}_{ij}=\Delta\mathbf{T}^{0}_{ij}$, its Jacobian with respect to a right perturbation is the identity, so the correction term contributes exactly $\mathbf{g}_{ij}$ to the gradient.
Together with the direct contribution from $\mathcal{U}$, $\tilde{\mathcal{U}}$ thus recovers the full gradient in \eqref{eq:chain} at the reference predictions, under the Gauss--Newton approximation and with the factor weights fixed.
As the predictions move away from the reference within a round, the correction provides a first-order approximation of the indirect contribution, which is refreshed with the updated network in the next round.
We therefore train the network by stochastic gradient descent on $\tilde{\mathcal{U}}$, so that both contributions propagate through the differentiable integration without unrolling BA.
This accounts for the network's influence on its geometric supervision, while letting BA retain its robust outlier handling.

\section{Visual-Inertial-Leg Localization}\label{sec:online}

\begin{figure*}[t]
  \centering
  \setlength{\abovecaptionskip}{-0.01cm}
  \includegraphics[width= 0.92\linewidth]{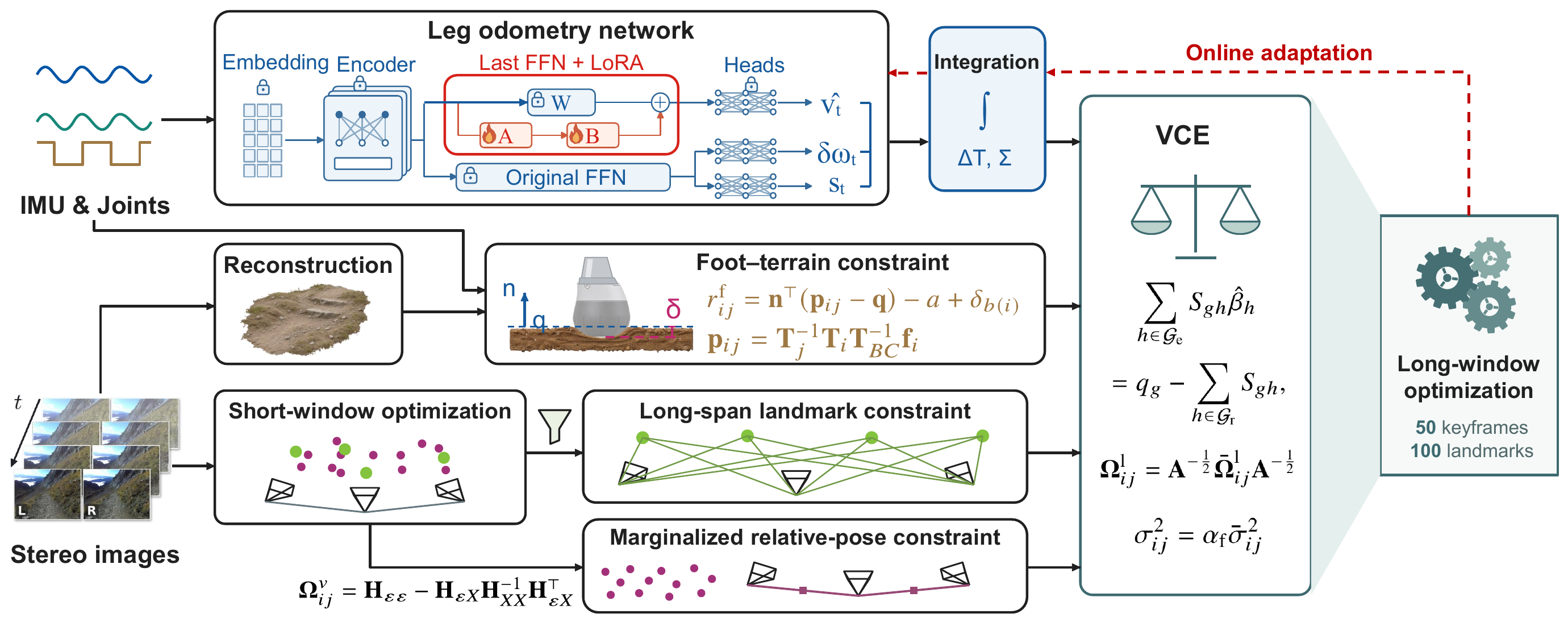}
  \caption{Overview of the online visual-inertial-leg localization system.
  The leg odometry network converts the IMU and joint measurements into relative motions between keyframes to aid the short-window stereo VO.
  The long-window fusion module combines the visual relative-pose, long-span landmark, leg odometry, foot-terrain, and inertial factors, with their weights calibrated by online VCE.
  The matured fused estimates supervise an online update of the low-rank adapters of the network, and the estimated gyroscope bias is fed back to the leg odometry.}
  \label{fig:vil_overview}
  \vspace{-1.0em}
\end{figure*}

\subsection{System Design}\label{sec:online-design}\label{sec:online-est}

In this section, we integrate the leg odometry network adapted in Section~\ref{sec:sim2real} into an online VIL localization system, as shown in Fig.~\ref{fig:vil_overview}.
The central difficulty in designing such a system is that the information carried by each sensor does not match the cost of exploiting it.
Visual measurements are rich, but each keyframe involves many map points, so they can be optimized only over a short window.
Proprioceptive measurements are inexpensive, but they relate only consecutive keyframes, so their errors accumulate over time.
We therefore adopt a dual-window design.
A stereo VO aided by the leg odometry optimizes a short window of keyframes with all their map points, whereas a long-window fusion module retains only a few high-quality landmarks, marginalizes the remaining points into relative-pose factors, and fuses them with the proprioceptive factors over a much longer window.

Furthermore, to reduce the vertical drift of the VIL system, we design a deformation-aware foot-terrain constraint that couples the leg kinematics with the vision.
Specifically, we use the vision information to reconstruct the terrain, and the joint measurements place each loaded foot on it, so every contact ties the current pose to a surface observed several keyframes earlier, at the cost of only a scalar residual.
A slowly varying contact offset absorbs the sinkage of the feet, keeping the constraint valid on deformable ground.
With this constraint, the fusion module combines visual, proprioceptive, and foot-terrain factors, whose reliability differs and changes with the terrain, motion, and visual conditions.
Instead of fixing their weights at the offline calibration of Section~\ref{sec:sim2real}, we re-estimate them by VCE whenever a window converges, using the visual factors as a reference and damping the updates across windows.

Overall, the fusion module estimates the poses of the $N$ most recent keyframes, together with the gyroscope biases of the leg odometry, the inertial states, the selected landmarks, and the contact offsets, by minimizing
\begin{equation}
  \begin{aligned}
  E={}&\sum_{(i,j)}\|\mathbf{r}_{ij}\|^2_{\bm{\Omega}^{\mathrm{l}}_{ij}}
  +E_{\mathrm{imu}}
  +\sum_{(i,j)}\|\mathbf{r}^{\mathrm{v}}_{ij}\|^2_{\bm{\Omega}^{\mathrm{v}}_{ij}}\\
  &+\sum_{(i,l)}\|\mathbf{r}^{\mathrm{p}}_{il}\|^2_{\bm{\Omega}^{\mathrm{p}}}
  +\sum_{(i,j)}\frac{(r^{\mathrm{f}}_{ij})^2}{\sigma_{ij}^2}
  +E_{\mathrm{prior}},
  \end{aligned}
  \label{eq:online_backend}
\end{equation}
where the first sum and $E_{\mathrm{imu}}$ are the proprioceptive factors (Section~\ref{sec:online-proprio}), namely the leg odometry factors and the inertial and zero-motion factors, the next two sums are the visual relative-pose and landmark reprojection factors (Section~\ref{sec:online-visual}), the last sum contains the foot-terrain factors (Section~\ref{sec:online-foot}), and $E_{\mathrm{prior}}$ contains the priors and random walks of the bias and offset blocks.
The weights $\bm{\Omega}^{\mathrm{l}}_{ij}$ and $\sigma_{ij}^2$ of the leg odometry and foot-terrain factors are rescaled online by VCE (Section~\ref{sec:online-vce}).

Although the leg odometry network has been adapted to the real robot, its accuracy may still degrade on terrains and ground conditions absent from its training data.
We therefore adapt the network online during deployment, using the fused estimates as supervision (Section~\ref{sec:online-adapt}).
The network runs in its own process, referred to as the network process.
To keep the update lightweight and preserve the priors learned in the previous stages, we train only a small set of low-rank adapters in this process, supervise them with poses that have been refined over several windows, and accept an update only if it does not degrade the predictions on held-out keyframe pairs.

\subsection{Proprioceptive Motion Factors}\label{sec:online-proprio}

The IMU and joint encoders measure the body motion at a high rate, independently of the visual conditions.
We incorporate these proprioceptive measurements through three types of factors: leg odometry factors that constrain the relative motion between keyframes, inertial factors that additionally constrain velocity and the gravity direction, and zero-motion factors that hold the pose fixed when the robot stands still.

\subsubsection{Leg Odometry Factors}\label{sec:online-leg}

We incorporate the network predictions into the estimator as relative-pose factors between keyframes.
For each keyframe pair $(i,j)$ fully covered by network predictions, the camera-frame motion $\Delta\mathbf{T}_{ij}$ integrated by \eqref{eq:integration} defines the residual
\begin{equation}
  \mathbf{r}_{ij}=\mathrm{Log}\big(\Delta\mathbf{T}_{ij}^{-1}\mathbf{T}_i^{-1}\mathbf{T}_j\big),
  \label{eq:online_leg}
\end{equation}
as in \eqref{eq:lower}, with a diagonal information matrix $\bar{\bm{\Omega}}^{\mathrm{l}}_{ij}$ calibrated offline and rescaled online by VCE (Section~\ref{sec:online-vce}).

However, the learned correction $\delta\bm{\omega}_k$ compensates only for motion-dependent errors, such as transients induced by foot impacts and vibration.
It cannot capture the gyroscope bias, which varies between power cycles and drifts slowly independently of the motion, and would thus accumulate into rotational drift.
We therefore estimate the bias explicitly in the fusion module, where the rotation is observed through visual, inertial, and leg constraints, and feed it back to the network process.
At start-up, $\mathbf{b}_g$ is initialized as $\bar{\mathbf{b}}_g$, estimated from the gyroscope readings over static intervals.
The fusion module then estimates the gyroscope bias $\mathbf{b}_m$, shared within each block $m$ of $N_g$ consecutive keyframes.
Since $\mathbf{b}_m$ represents the entire bias rather than a residual of $\mathbf{b}_g$, the leg rotation $\Delta\mathbf{R}^{B}_{ij}$ is integrated from $\bm{\omega}^{\mathrm{m}}_k+\delta\bm{\omega}_k$ without subtracting $\mathbf{b}_g$, and the leg odometry is corrected as
\begin{equation}
  \Delta\mathbf{T}^{B}_{ij}(\mathbf{b}_{m})=
  \big(
    \Delta\mathbf{R}^{B}_{ij}\,
    \mathrm{Exp}(-\mathbf{b}_{m}\,\Delta t_{ij}),
    \ \Delta\mathbf{p}^{B}_{ij}
  \big),
  \label{eq:online_gyro_bias}
\end{equation}
where $\Delta t_{ij}=t_j-t_i$ and $m=\lfloor j/N_g\rfloor$ is the block of keyframe $j$.
This correction is a first-order approximation over the short keyframe interval.
The biases are regularized in $E_{\mathrm{prior}}$ by
\begin{equation}
  E_{g}=\frac{\|\mathbf{b}_0-\bar{\mathbf{b}}_g\|^2}{\sigma_{g}^2}+\sum_{m\geq1}\frac{\|\mathbf{b}_m-\mathbf{b}_{m-1}\|^2}{\sigma_{g,\mathrm{rw}}^2},
  \label{eq:online_bias_prior}
\end{equation}
where the prior on the first block propagates to later blocks through the random walk, allowing the bias to vary gradually over time.
After each solve, the newest block replaces $\mathbf{b}_g$ in the network process, so that the network inputs are compensated by the bias estimated from all constraints.

\subsubsection{Inertial Factors}\label{sec:online-imu}
Beyond the relative motion predicted by the leg odometry, the raw IMU measurements constrain roll and pitch through the gravity direction.
We therefore add IMU preintegration factors~\cite{forster2016manifold} between consecutive keyframes, which introduce per-keyframe velocities, bias blocks with random-walk priors, and the gravity direction as additional states.
To compensate for small camera-IMU rotation errors, which project gravity onto the wrong axes and cause drift, we also estimate a rotation correction $\bm{\psi}$ of the IMU frame, shared by the window and regularized by a zero-mean prior.
Specifically, let $\mathbf{R}_i$, $\mathbf{p}_i$, and $\mathbf{v}_i$ denote the orientation, position, and velocity of the IMU at keyframe $i$, obtained from $\mathbf{T}_i$ through the calibrated extrinsic.
With the corrected orientation $\mathbf{R}'_i=\mathbf{R}_i\mathrm{Exp}(\bm{\psi})$, the preintegration residuals become
\begin{equation}
  \begin{aligned}
  \mathbf{r}^{\Delta R}_{ij}&=\mathrm{Log}\big(\Delta\tilde{\mathbf{R}}_{ij}^{\top}\mathbf{R}'^{\top}_i\mathbf{R}'_j\big),\\
  \mathbf{r}^{\Delta v}_{ij}&=\mathbf{R}'^{\top}_i\big(\mathbf{v}_j-\mathbf{v}_i-\mathbf{g}_w\Delta t_{ij}\big)-\Delta\tilde{\mathbf{v}}_{ij},\\
  \mathbf{r}^{\Delta p}_{ij}&=\mathbf{R}'^{\top}_i\big(\mathbf{p}_j-\mathbf{p}_i-\mathbf{v}_i\Delta t_{ij}-\tfrac{1}{2}\mathbf{g}_w\Delta t_{ij}^2\big)-\Delta\tilde{\mathbf{p}}_{ij},
  \end{aligned}
  \label{eq:online_imu}
\end{equation}
where $\Delta\tilde{\mathbf{R}}_{ij}$, $\Delta\tilde{\mathbf{v}}_{ij}$, and $\Delta\tilde{\mathbf{p}}_{ij}$ are the preintegrated measurements, corrected to first order for the current bias estimates~\cite{forster2016manifold}, and $\mathbf{g}_w$ is the gravity vector in the world frame, whose direction is estimated under a prior.
The residuals are weighted by the inverse of the preintegration covariance.

\subsubsection{Zero-Motion Factors}\label{sec:online-zupt}
When the robot stands still, we use a zero-motion constraint to suppress drift.
To avoid false detections, we declare an interval $[t_i,t_j]$ stationary only when the IMU and the leg odometry agree, that is, when
\begin{equation}
  \max_k\|\tilde{\bm{\omega}}_k\|<\omega_{\mathrm{th}},\quad
  \big\|\mathrm{std}_k(\tilde{\mathbf{a}}_k)\big\|_{\infty}<a_{\mathrm{th}},\quad
  \|\Delta\mathbf{p}_{ij}\|<p_{\mathrm{th}},
  \label{eq:online_zupt_detect}
\end{equation}
where $\tilde{\bm{\omega}}_k$ and $\tilde{\mathbf{a}}_k$ are the gyroscope and accelerometer samples in the interval, $\mathrm{std}_k(\cdot)$ is their per-axis standard deviation, $\Delta\mathbf{p}_{ij}$ is the leg odometry translation, and $\omega_{\mathrm{th}}$, $a_{\mathrm{th}}$, and $p_{\mathrm{th}}$ are thresholds.
For each such interval, we add a zero-motion factor
\begin{equation}
  \mathbf{r}^{z}_{ij}=\mathrm{Log}\big(\mathbf{T}_i^{-1}\mathbf{T}_j\big),\qquad
  \bm{\Omega}^{z}=\mathrm{diag}\big(\sigma_{r}^{-2}\mathbf{I}_3,\ \sigma_{t}^{-2}\mathbf{I}_3\big),
  \label{eq:online_zupt}
\end{equation}
where the small standard deviations $\sigma_r$ and $\sigma_t$ hold the pose fixed during standing and thus suppress drift.
In addition, the network process detects standing causally, when all joint speeds and the per-axis gyroscope standard deviations over the preceding $0.5$\,s stay below thresholds, and outputs a zero twist for these samples, so that the leg odometry factors also hold the pose fixed while the robot stands still.

\subsection{Lightweight Long-Window Visual Factors}\label{sec:online-visual}\label{sec:online-optional}

Visual measurements enter the estimator at two levels.
In the short window, a leg-aided stereo VO optimizes a few recent keyframes together with all map points observed in them.
In the long window, since optimizing all map points would be costly, we select a small number of high-quality map points as landmarks and constrain them by their stereo reprojection errors.
The remaining map points are marginalized and enter the optimization as relative-pose factors between consecutive keyframes, which keeps their constraints on the relative motion at a low cost.
We next describe these components.

\subsubsection{Long-Span Landmark Factors}\label{sec:online-landmark}
We select at most $K=100$ landmarks that are observed in stereo by at least three keyframes of the window, ranked by their track length and temporal span.
To distribute them over the scene, we apply non-maximum suppression in 3D space, which keeps only the highest-ranked landmark within each local region.
Landmarks selected in the previous solve are retained first if they remain valid, which keeps the estimates of consecutive overlapping windows consistent.
For an observation $\tilde{\mathbf{x}}_{il}=[u,v,u_r]^{\top}$ of landmark $l$ in keyframe $i$, the stereo reprojection residual is
\begin{equation}
  \mathbf{r}^{\mathrm{p}}_{il}=\tilde{\mathbf{x}}_{il}-
  \begin{bmatrix}
    f_xX/Z+c_x\\
    f_yY/Z+c_y\\
    f_xX/Z+c_x-f_xb/Z
  \end{bmatrix},
  \label{eq:online_landmark}
\end{equation}
where $(u,v)$ and $u_r$ are the pixel coordinates in the left image and the horizontal coordinate in the right image, $[X,Y,Z]^{\top}=\mathbf{T}_i^{-1}\mathbf{X}_l$ is the landmark in the camera frame, $f_x$, $f_y$, $c_x$, and $c_y$ are the camera intrinsics, and $b$ is the stereo baseline.

\subsubsection{Map Point Marginalization}\label{sec:online-marg}
The map points that are not selected as landmarks still carry substantial visual information.
Rather than discarding them, we marginalize them onto each pair of consecutive keyframes $(i,j)$, which yields a compact relative-pose constraint without enlarging the state of the long window.
Specifically, let $\mathcal{P}_{ij}$ denote the unselected map points observed by both keyframes.
We linearize their stereo reprojection residuals $\mathbf{r}^{\mathrm{p}}_{kl}$, $k\in\{i,j\}$, as in \eqref{eq:online_landmark}, at the VO estimates of the poses and points.
To remove the gauge freedom of the pair, $\mathbf{T}_i$ is held fixed, and $\mathbf{T}_j$ and the points are perturbed as $\mathbf{T}_j\leftarrow\mathbf{T}_j\mathrm{Exp}(\bm{\varepsilon})$ and $\mathbf{X}_l\leftarrow\mathbf{X}_l+\delta\mathbf{X}_l$.
For each point $l\in\mathcal{P}_{ij}$, the Gauss--Newton Hessian consists of the blocks
\begin{equation}
  \begin{aligned}
  \mathbf{H}^{\varepsilon\varepsilon}_l&=\gamma_{jl}(\mathbf{J}^{\varepsilon}_{jl})^{\top}\mathbf{J}^{\varepsilon}_{jl},\qquad
  \mathbf{H}^{\varepsilon X}_l=\gamma_{jl}(\mathbf{J}^{\varepsilon}_{jl})^{\top}\mathbf{J}^{X}_{jl},\\
  \mathbf{H}^{XX}_l&=\sum_{k\in\{i,j\}}\gamma_{kl}(\mathbf{J}^{X}_{kl})^{\top}\mathbf{J}^{X}_{kl},
  \end{aligned}
  \label{eq:online_marg_blocks}
\end{equation}
where $\mathbf{J}^{\varepsilon}_{kl}$ and $\mathbf{J}^{X}_{kl}$ are the Jacobians of $\mathbf{r}^{\mathrm{p}}_{kl}$ with respect to $\bm{\varepsilon}$ and $\mathbf{X}_l$, whitened by the pixel noise $\sigma_{\mathrm{px}}$, and $\gamma_{kl}=\min\big(1,\kappa\,\sigma_{\mathrm{px}}/\|\mathbf{r}^{\mathrm{p}}_{kl}\|\big)$ is the Huber weight with threshold $\kappa$.
Since $\mathbf{T}_i$ is fixed, the observation in keyframe $i$ contributes only to the point block $\mathbf{H}^{XX}_l$ and does not enter the pose blocks.

Stacking all points of $\mathcal{P}_{ij}$ gives the pairwise Hessian
\begin{equation}
  \mathbf{H}_{ij}=
  \begin{bmatrix}
    \mathbf{H}_{\varepsilon\varepsilon} & \mathbf{H}_{\varepsilon X}\\
    \mathbf{H}_{\varepsilon X}^{\top} & \mathbf{H}_{XX}
  \end{bmatrix},
  \label{eq:online_marg_hessian}
\end{equation}
where $\mathbf{H}_{\varepsilon\varepsilon}=\sum_{l}\mathbf{H}^{\varepsilon\varepsilon}_l$, $\mathbf{H}_{\varepsilon X}$ concatenates the pose-point blocks $\mathbf{H}^{\varepsilon X}_l$, and $\mathbf{H}_{XX}=\mathrm{diag}(\mathbf{H}^{XX}_l)$ is block diagonal.
Marginalizing the points by the Schur complement yields the information matrix of the relative pose,
\begin{equation}
  \begin{aligned}
  \bm{\Omega}^{\mathrm{v}}_{ij}
  &=\mathbf{H}_{\varepsilon\varepsilon}-\mathbf{H}_{\varepsilon X}\mathbf{H}_{XX}^{-1}\mathbf{H}_{\varepsilon X}^{\top}\\
  &=\sum_{l\in\mathcal{P}_{ij}}\Big(\mathbf{H}^{\varepsilon\varepsilon}_l-\mathbf{H}^{\varepsilon X}_l\big(\mathbf{H}^{XX}_l\big)^{-1}\big(\mathbf{H}^{\varepsilon X}_l\big)^{\top}\Big).
  \end{aligned}
  \label{eq:online_schur}
\end{equation}
Because $\mathbf{H}_{XX}$ is block diagonal, the Schur complement reduces to independent $3\times3$ inversions, and its cost grows only linearly with the number of points.
We use this information matrix to weight a relative-pose factor centered at the VO estimate $\Delta\mathbf{T}^{\mathrm{v}}_{ij}=(\mathbf{T}^{\mathrm{vo}}_i)^{-1}\mathbf{T}^{\mathrm{vo}}_j$, refined by local BA:
\begin{equation}
  \mathbf{r}^{\mathrm{v}}_{ij}=\mathrm{Log}\big((\Delta\mathbf{T}^{\mathrm{v}}_{ij})^{-1}\mathbf{T}_i^{-1}\mathbf{T}_j\big).
  \label{eq:online_visual}
\end{equation}
The factor enters \eqref{eq:online_backend} with weight $\bm{\Omega}^{\mathrm{v}}_{ij}$.
It is an approximation that neglects the remaining gradient at the VO estimate, the correlation between pairs that share map points, and the fact that the VO estimate already depends on the leg odometry.
Its full information matrix preserves the directional constraints induced by the feature geometry and the coupling between rotation and translation, while requiring only one six-dimensional factor per pair.
For numerical stability and robustness, we exclude observations behind the camera or with excessive reprojection errors, as well as points with nearly singular $\mathbf{H}^{XX}_l$,
and no factor is formed if too few valid points remain.

\subsection{Deformation-Aware Foot-Terrain Constraint}\label{sec:online-foot}

\begin{figure}[t]
  \centering
  \setlength{\abovecaptionskip}{-0.3cm}
  \includegraphics[width=\columnwidth]{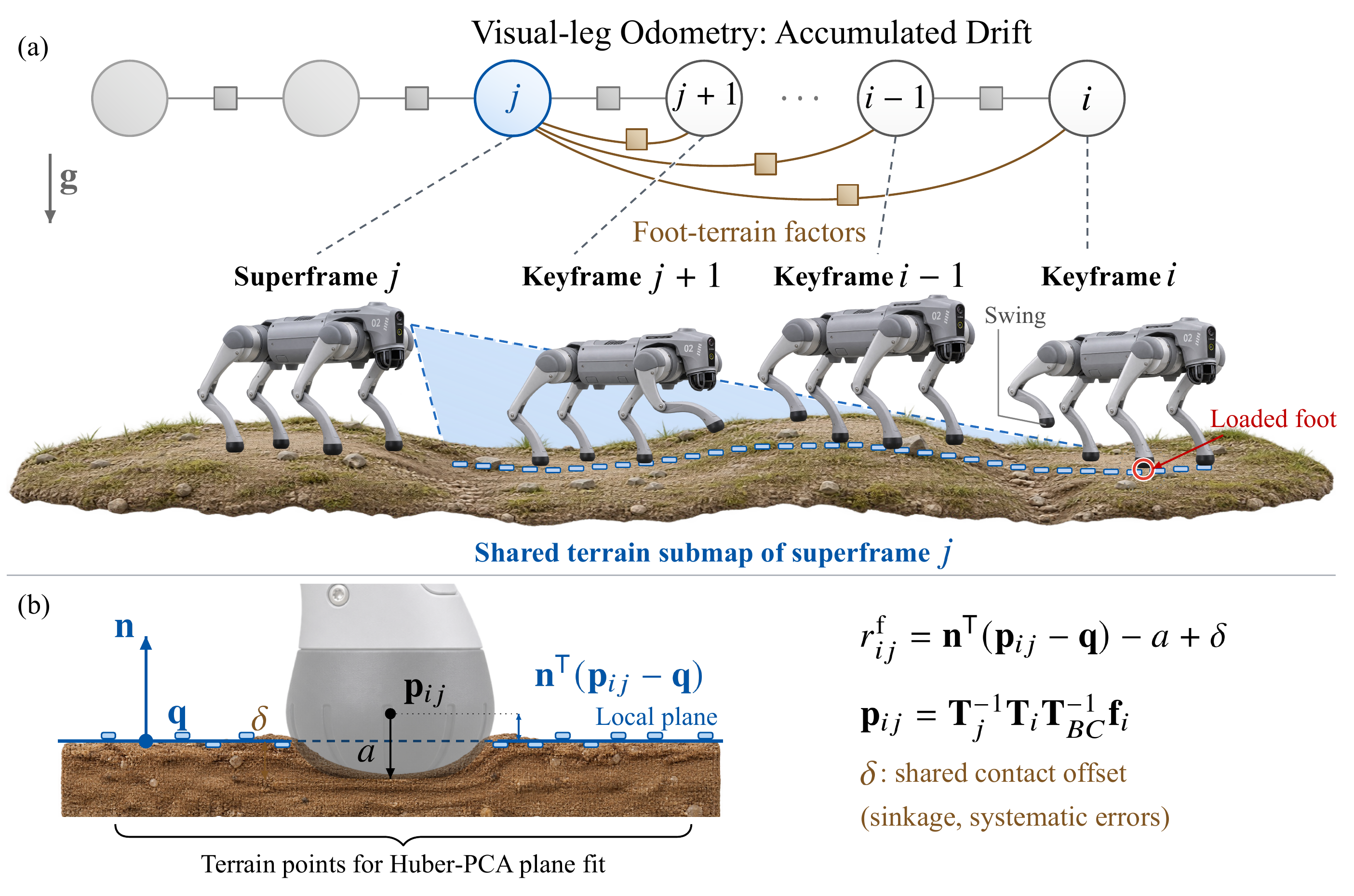}
  \caption{Foot-terrain constraint.
  (a) A loaded foot at keyframe $i$ rests on the terrain reconstructed earlier from superframe $j$, so the foot-terrain factor relates $\mathbf{T}_i$ directly to $\mathbf{T}_j$, bypassing the drift accumulated along the keyframe chain.
  (b) The residual is the signed distance between the foot center $\mathbf{p}_{ij}$ and a local plane $(\mathbf{q},\mathbf{n})$, corrected by the foot radius $a$ and the contact offset $\delta$.}
  \label{fig:foot_terrain}
  \vspace{-1.em}
\end{figure}

Body vibrations during legged locomotion introduce additional disturbances into the onboard accelerometer measurements, which can exacerbate vertical drift in VIO and VIL estimation.
To mitigate this drift, we design and incorporate a foot-terrain constraint into the optimization framework.
Our key insight is that a loaded foot is supported by the terrain, physically linking the robot pose to the supporting surface.
Combining this contact information with visually reconstructed terrain therefore provides an additional constraint on the robot's relative height.
However, enforcing exact contact with the reconstructed surface can introduce estimation errors on deformable ground, where loaded feet may sink.
We therefore model this discrepancy as a contact offset, assuming that sinkage varies slowly over a short temporal window.
This formulation accommodates foot sinkage and modeling errors while retaining a geometric constraint on relative height, helping reduce vertical drift as illustrated in Fig.~\ref{fig:foot_terrain}.

\subsubsection{Superframe-Based Terrain Submaps}

As the foot-terrain factors require local surface geometry that the sparse VO map cannot adequately provide, we construct dense terrain submaps by estimating stereo depth with Fast-FoundationStereo~\cite{wen2026fastfoundationstereo} and back-projecting it into 3D.
To reduce the computational cost of dense stereo estimation, we reconstruct terrain only at a sparse subset of keyframes, termed \textit{superframes}.
We select these superframes to maintain sufficient terrain coverage while minimizing redundant reconstruction.
Specifically, a keyframe becomes a superframe when the translation or rotation since the previous superframe exceeds a threshold, or when existing submaps cover fewer than half of the current stance feet.
Each submap is anchored to its superframe, allowing pose updates to reposition the terrain without repeating reconstruction.

\subsubsection{Contact-Surface Association}

Reliable foot-terrain factors require both stable contact and correct surface association; otherwise, they can bias the estimated height toward an incorrect surface.
We therefore screen contacts and candidate planes conservatively, maintain a fixed association within each contact segment, and weight accepted factors by their estimated uncertainty.
In practice, we first identify contact segments from the normalized foot forces in \eqref{eq:force_norm}, using hysteresis thresholds to suppress rapid switching between contact states.
Within each segment, which begins at touchdown and ends at lift-off, we exclude observations near both transitions to retain only stable stance measurements.
For each retained observation at keyframe $i$, we then compute the body-frame foot position $\mathbf{f}_i$ from joint measurements via forward kinematics.

To identify the supporting surface, we retrieve terrain points $\{\mathbf{x}_k\}$ around the foot from the latest submaps and fit a plane by Huber-weighted principal component analysis, iterating
\begin{equation}
  \begin{gathered}
  \mathbf{q}=\frac{\sum_k w_k\mathbf{x}_k}{\sum_k w_k},\qquad
  \mathbf{M}=\frac{\sum_k w_k(\mathbf{x}_k-\mathbf{q})(\mathbf{x}_k-\mathbf{q})^{\top}}{\sum_k w_k},\\
  w_k=\min\Big(1,\ \frac{\epsilon_{\mathrm{p}}}{|\mathbf{n}^{\top}(\mathbf{x}_k-\mathbf{q})|}\Big),
  \end{gathered}
  \label{eq:foot_plane}
\end{equation}
where the unit normal $\mathbf{n}$ is the eigenvector of $\mathbf{M}$ associated with its smallest eigenvalue, and $\epsilon_{\mathrm{p}}$ is the Huber threshold.
The plane is represented by $\mathbf{q}$ and $\mathbf{n}$ in the camera frame of superframe $j$.
To avoid associating the foot with an ambiguous or poorly reconstructed surface, we reject fits with stepped or multilayered height distributions, incomplete support around the foot, poor planarity, indicated by a large ratio $\lambda_1/(\lambda_1+\lambda_2+\lambda_3)$ of the ascending eigenvalues of $\mathbf{M}$, or excessive fitting error.

Since a stance foot should remain on the same surface throughout a contact segment, we keep its plane association fixed and discard the segment if subsequent observations become inconsistent.
To give less weight to distant or poorly fitted surfaces, we model the factor variance as
\begin{equation}
  \sigma_{ij}^2=\sigma_0^2+(c\,z^2)^2+e_{\mathrm{rms}}^2,
  \label{eq:foot_var}
\end{equation}
where $\sigma_0^2$ is a baseline variance, $c\,z^2$ captures the quadratic growth of stereo depth error with surface depth $z$, and $e_{\mathrm{rms}}$ is the root-mean-square plane-fitting error.

\subsubsection{Foot-Terrain Factors}

Each accepted association introduces a foot-terrain factor linking keyframe $i$ to the terrain anchored at superframe $j$ (Fig.~\ref{fig:foot_terrain}a).
By enforcing consistency between the kinematic foot position and the visually reconstructed supporting surface, this factor turns terrain geometry into a direct constraint on robot-to-terrain height, helping suppress vertical drift accumulated through successive relative-motion estimates.
Specifically, we transform the foot position into the camera frame of superframe $j$ and project its displacement from the supporting plane onto the normal $\mathbf{n}$.
Accounting for the foot radius $a$ and contact offset $\delta$ gives
\begin{equation}
  r_{ij}^{\mathrm{f}}
  =\mathbf{n}^{\top}
  \big(
    \mathbf{T}_j^{-1}\mathbf{T}_i
    \mathbf{T}_{BC}^{-1}\mathbf{f}_i-\mathbf{q}
  \big)-a+\delta,
  \label{eq:online_foot}
\end{equation}
with the residual weighted by the inverse variance $1/\sigma_{ij}^2$ defined in \eqref{eq:foot_var}.
This residual constrains motion along the plane normal, while tangential motion is governed by the other factors.

In practice, the offset $\delta$ absorbs not only foot sinkage but also systematic discrepancies from kinematic, extrinsic, and depth errors.
However, a single factor cannot distinguish $\delta$ from the height of keyframe $i$ along the plane normal:
\begin{equation}
  r^{\mathrm{f}}_{ij}\big(\mathbf{T}_i\oplus\Delta h\,\mathbf{n},\ \delta-\Delta h\big)=r^{\mathrm{f}}_{ij}\big(\mathbf{T}_i,\ \delta\big),
  \label{eq:foot_ambiguity}
\end{equation}
where $\mathbf{T}_i\oplus\Delta h\,\mathbf{n}$ denotes translating keyframe $i$ by $\Delta h$ along $\mathbf{n}$.
That is, lowering the robot relative to the terrain can be compensated by an equal increase in $\delta$, so an independent offset for each contact would absorb the very height errors that the factor is intended to correct.
We therefore share one offset $\delta_b$ among all contacts within each block $b$ of $N_b$ consecutive keyframes, i.e., $\delta$ in \eqref{eq:online_foot} is replaced by $\delta_{b(i)}$ with $b(i)=\lfloor i/N_b\rfloor$.
Each $\delta_b$ is bounded to a physically plausible range $[\delta_{\min},\delta_{\max}]$, and the offsets are regularized by the following term, which is included in $E_{\mathrm{prior}}$ of \eqref{eq:online_backend}:
\begin{equation}
  E_{\delta}=\sum_{b}\frac{\delta_b^2}{\sigma_{\delta}^2}+\sum_{b}\frac{(\delta_b-\delta_{b-1})^2}{\sigma_{\mathrm{rw}}^2},
  \label{eq:foot_offset_prior}
\end{equation}
where $\sigma_{\delta}$ and $\sigma_{\mathrm{rw}}$ are the standard deviations of the zero-mean prior and of the random walk between adjacent blocks, so that the offset varies only slowly over time.
This design allows the offset to track gradual changes in ground conditions while retaining sensitivity to short-term vertical errors.

\subsection{Online Variance Component Estimation}\label{sec:online-vce}

As terrain, motion, and visual conditions vary, the offline calibration in Section~\ref{sec:sim2real} may no longer reflect the relative uncertainty of the factors in \eqref{eq:online_backend}.
We therefore apply the VCE formulation in Section~\ref{sec:s2-lower} to each converged window, adjusting the calibrated weights according to the observed residual statistics.
We estimate four variance scales: $\alpha_{\mathrm{R}}$, $\alpha_{xy}$, and $\alpha_z$ for the rotational, horizontal translational, and vertical translational components of leg odometry, respectively, and $\alpha_{\mathrm{f}}$ for the foot-terrain factors.
These scales modify the calibrated information matrix $\bar{\bm{\Omega}}^{\mathrm{l}}_{ij}$ and variance $\bar{\sigma}_{ij}^2$ as
\begin{equation}
  \begin{gathered}
  \bm{\Omega}^{\mathrm{l}}_{ij}=\mathbf{D}^{-\frac{1}{2}}\bar{\bm{\Omega}}^{\mathrm{l}}_{ij}\mathbf{D}^{-\frac{1}{2}},\qquad
  \sigma^2_{ij}=\alpha_{\mathrm{f}}\bar{\sigma}^2_{ij},\\
  \mathbf{D}=\mathrm{diag}\big(\alpha_{\mathrm{R}}\mathbf{I}_3,\ \alpha_{xy}\mathbf{I}_2,\ \alpha_z\big).
  \end{gathered}
  \label{eq:online_vce_groups}
\end{equation}
Since the errors of leg odometry differ across directions and are typically largest in the vertical direction, separate scales allow each direction to be weighted independently.

However, the visual and leg odometry weights cannot be separated from the data alone.
Specifically, for a keyframe pair connected by one visual and one leg odometry factor, both measure the same relative motion, and to first order the data only reveal their discrepancy
\begin{equation}
  \mathrm{Log}\big(\Delta\mathbf{T}_{ij}^{-1}\Delta\mathbf{T}^{\mathrm{v}}_{ij}\big)\approx\mathbf{n}^{\mathrm{v}}_{ij}-\mathbf{n}^{\mathrm{l}}_{ij}\sim\mathcal{N}\big(\mathbf{0},\,\bm{\Sigma}^{\mathrm{v}}_{ij}+\bm{\Sigma}^{\mathrm{l}}_{ij}\big),
  \label{eq:online_vce_ident}
\end{equation}
which determines only the sum of the two covariances, not how it is divided between them.
We therefore keep the visual and prior factors as a reference set $\mathcal{G}_{\mathrm{r}}$ with $\beta_h=1$, and solve the moment equations \eqref{eq:vce} only for the estimated groups $\mathcal{G}_{\mathrm{e}}$:
\begin{equation}
  \sum_{h\in\mathcal{G}_{\mathrm{e}}}S_{gh}\hat{\beta}_h=q_g-\sum_{h\in\mathcal{G}_{\mathrm{r}}}S_{gh},\qquad g\in\mathcal{G}_{\mathrm{e}}.
  \label{eq:online_vce_solve}
\end{equation}
A group is moved to $\mathcal{G}_{\mathrm{r}}$ if its redundancy \eqref{eq:redundancy} falls below a threshold or its estimate is not positive.
The required traces involve only selected entries of $\mathbf{H}^{-1}$, which we compute from its sparse factorization.
Since consecutive windows overlap, we perform one damped update \eqref{eq:vce_update} per window and carry the estimates across windows, with the scales bounded to $[\alpha_{\min},\alpha_{\max}]$ so that the weights do not drift far from their calibrated values.
The window is then optimized again with the updated weights, so that the scales converge gradually over successive windows at a small additional cost.

\subsection{Online Leg Odometry Adaptation}\label{sec:online-adapt}

During deployment, the leg odometry network may encounter conditions that were not represented in its adaptation data, such as new terrains.
To handle such conditions, we fine-tune the network online, using the fused estimates as supervision.
To keep this update lightweight, we add low-rank adapters~\cite{hu2022lora} to the network and train only these.
Specifically, the weight $\mathbf{W}$ of each linear layer in the feed-forward block of the last Transformer layer is replaced by
\begin{equation}
  \mathbf{W}'=\mathbf{W}+\tfrac{\alpha}{r}\,\mathbf{B}\mathbf{A},
  \label{eq:online_lora}
\end{equation}
where $\mathbf{A}$ and $\mathbf{B}$ have rank $r=1$, and all original parameters are frozen.
Since $\mathbf{B}$ is initialized to zero, the network initially produces the same predictions as before online adaptation, and only about $1.3$k parameters are trained.
The adapted feature is passed only to the linear-velocity head, whereas the gyroscope-correction and uncertainty heads read the feature of the original layer, so the adapters change only the predicted linear velocity and leave $\delta\bm{\omega}_k$ and the predicted uncertainties unchanged.

Unlike Section~\ref{sec:sim2real}, where the network is supervised through the reprojection errors of 3D map points, we supervise the online adapters directly with the fused poses of consecutive keyframe pairs $(i,j)\in\mathcal{P}$ for better online efficiency.
For each pair, the network predictions are integrated by \eqref{eq:integration} into $\Delta\mathbf{T}^{B}_{ij}$ and compared with the fused motion in the body frame:
\begin{equation}
  \begin{gathered}
  \Delta\mathbf{T}^{B,*}_{ij}=\mathbf{T}_{BC}\,\mathbf{T}_i^{-1}\mathbf{T}_j\,\mathbf{T}_{BC}^{-1},\\
  \big[(\mathbf{e}^{R}_{ij})^{\top},(\mathbf{e}^{t}_{ij})^{\top}\big]^{\top}=\mathrm{Log}\big((\Delta\mathbf{T}^{B,*}_{ij})^{-1}\Delta\mathbf{T}^{B}_{ij}\big).
  \end{gathered}
  \label{eq:online_adapt_pose}
\end{equation}
To regularize the predicted velocities between keyframes, we also compare their change with the IMU measurements:
\begin{equation}
  \mathbf{r}^{a}_{ij}=\Delta\mathbf{R}_{ij}\hat{\mathbf{v}}^{I}_{j}-\hat{\mathbf{v}}^{I}_{i}-\sum_k\big(\Delta\mathbf{R}_{ik}\tilde{\mathbf{a}}_k+\mathbf{g}^{B}_i\big)\Delta t_k,
  \label{eq:online_adapt_acc}
\end{equation}
where $\hat{\mathbf{v}}^{I}_{i}$ and $\hat{\mathbf{v}}^{I}_{j}$ are the first and last predicted velocities of the interval, transferred to the IMU location, $\Delta\mathbf{R}_{ik}$ is the integrated rotation in \eqref{eq:integration}, and $\mathbf{g}^{B}_i$ is the gravity vector expressed in the body frame at $t_i$ using the fused orientation.
The adapter parameters $\bm{\theta}_{\mathrm{L}}$ are trained by minimizing
\begin{equation}
  \begin{aligned}
  \mathcal{L}={}&\frac{1}{|\mathcal{P}|}\!\sum_{(i,j)\in\mathcal{P}}\!\big(\rho(\mathbf{e}^{t}_{ij})+\lambda_{a}\rho(\mathbf{r}^{a}_{ij})\big)
  +\lambda_{s}\mathcal{L}_{s}+\lambda_{0}\big\|\bm{\theta}_{\mathrm{L}}-\bm{\theta}^{0}_{\mathrm{L}}\big\|^2,
  \end{aligned}
  \label{eq:online_adapt_loss}
\end{equation}
where $\rho(\cdot)$ is the Huber function, $\mathcal{L}_{s}$ penalizes temporal changes of the predicted velocity, and the last term anchors the adapters to their initial values $\bm{\theta}^{0}_{\mathrm{L}}$.
Training is performed in rounds in the network process.
A round is triggered by every ten new keyframes in $\mathcal{P}$ and takes five AdamW steps on the pairs among the ten most recent of them, with every fifth pair held out.
The update is accepted only if it does not increase the data terms of $\mathcal{L}$ on the held-out pairs.
If the held-out loss of the adapted network exceeds that of the original network by more than $20\%$ for three consecutive rounds, the adapters are reset.
Accepted adapters replace the ones used for inference, and they affect only subsequent predictions.

\input{figures/sota_table}

\subsection{Global Map Refinement}\label{sec:online-global}

The online estimator operates on a sliding window and therefore cannot correct the drift accumulated beyond it.
After a run, we refine the estimates offline to build a globally consistent map.
This step builds upon the loop detection and global BA of AirSLAM~\cite{xu2025airslam}, but instead of the visual-only global BA, we construct the factor graph of \eqref{eq:online_backend} over the entire trajectory in a single window and add the loop-closure factor:
\begin{equation}
  E_{\mathrm{g}}=E+\sum_{(i,j)\in\mathcal{C}}\big\|\mathrm{Log}\big((\Delta\mathbf{T}^{\mathrm{lc}}_{ij})^{-1}\mathbf{T}_i^{-1}\mathbf{T}_j\big)\big\|^2_{\bm{\Omega}^{\mathrm{lc}}},
  \label{eq:global}
\end{equation}
where $E$ is the cost \eqref{eq:online_backend} evaluated over all keyframes, $\mathcal{C}$ is the set of accepted loops, $\Delta\mathbf{T}^{\mathrm{lc}}_{ij}$ is the relative pose estimated by the loop verification, and $\bm{\Omega}^{\mathrm{lc}}$ is a fixed diagonal information matrix.
In this way, the leg odometry and foot-terrain factors, which are absent from the original global BA, also constrain the globally consistent solution; in particular, the foot-terrain factors continue to limit vertical drift between loop closures.

\section{Experiments} \label{sec:experiments}

In this section, we evaluate the accuracy, robustness, and efficiency of AirLeg.
We first compare AirLeg with state-of-the-art methods of different sensor configurations on a large-scale public legged-robot dataset (Section~\ref{sec:exp-sota}) and on a low-cost quadruped with a different sensor setup (Section~\ref{sec:exp-go2}).
We then evaluate the loop closure and global refinement (Section~\ref{sec:exp-loop}), the generalization of the adaptation across scenes (Section~\ref{sec:exp-scene}) and its dependence on the amount of real-world data (Section~\ref{sec:exp-data-duration}), the contribution of each component (Section~\ref{sec:exp-ablation}), and finally the runtime of the complete system on a legged robot (Section~\ref{sec:exp-runtime}).

\subsection{Comparison with State-of-the-Art}\label{sec:exp-sota}

\subsubsection{Dataset}
We evaluate on the GrandTour dataset~\cite{tuna2026grandtour}, which was recorded by an ANYmal~D quadruped~\cite{hutter2016anymal} in 49 missions across alpine trails, forests, riverbanks, construction sites, industrial plants, railway tracks, and urban streets and buildings, under sunlight, clouds, rain, dusk, night, and indoor smoke.
The ground truth is the position of a prism on the robot tracked by a Leica MS60 total station.
Since the total station must keep the prism in sight, the publicly released ground truth covers only part of each mission.
We therefore select the test sequences by their ground-truth coverage.
After excluding four missions (SPX-1, SPX-2, ARC-1, and ARC-7) whose released images are heavily corrupted by anonymization, we rank the remaining 45 missions by coverage, keep only the best-covered mission among those recorded at the same site, and retain the ones with a coverage above 50\%.
This yields 17 test sequences, and the other 28 sequences are used for the self-supervised sim-to-real adaptation in Section~\ref{sec:sim2real} without their ground truth.
According to their scenes and terrains, we further divide the test sequences into 8 normal and 9 challenging ones (Table~\ref{tab:sota}).
The challenging sequences contain conditions that degrade vision (V), such as night, rain, and abrupt lighting changes, or terrains that degrade leg odometry (L), such as ice, mud, snow, and soft sand, where the feet slip or sink.

\subsubsection{Baselines and Metrics}
We compare AirLeg with 18 open-source methods of four types: leg odometry, stereo VO, VIO, and VIL odometry.
The leg odometry methods use the IMU and the joint states (400\,Hz), the VO methods use the front stereo pair of the Alphasense camera bundle (10\,Hz), the VIO methods use the stereo pair and the IMU, and the VIL methods use all of them, except that the two VIL methods requiring depth images use the ZED2i RGB-D camera instead of the stereo pair.
For AirLeg, the IMU and joint states are downsampled to 100\,Hz as the input of the leg odometry network.
Importantly, unlike the benchmark in the original GrandTour paper~\cite{tuna2026grandtour}, which uses an expensive tactical-grade IMU, all methods, including AirLeg, use only the built-in IMU of ANYmal~D, as available on typical legged robots, which makes accurate state estimation considerably more challenging.
For a fair comparison, loop closure detection and correction are disabled in all methods.
We report the RMSE of the absolute trajectory error (ATE) after SE(3) alignment.
AirLeg is evaluated on its final back-end trajectory (two-run mean).

\begin{figure*}[!t]
  \centering
  \setlength{\abovecaptionskip}{-0.05cm}
  \includegraphics[width= 0.97\linewidth]{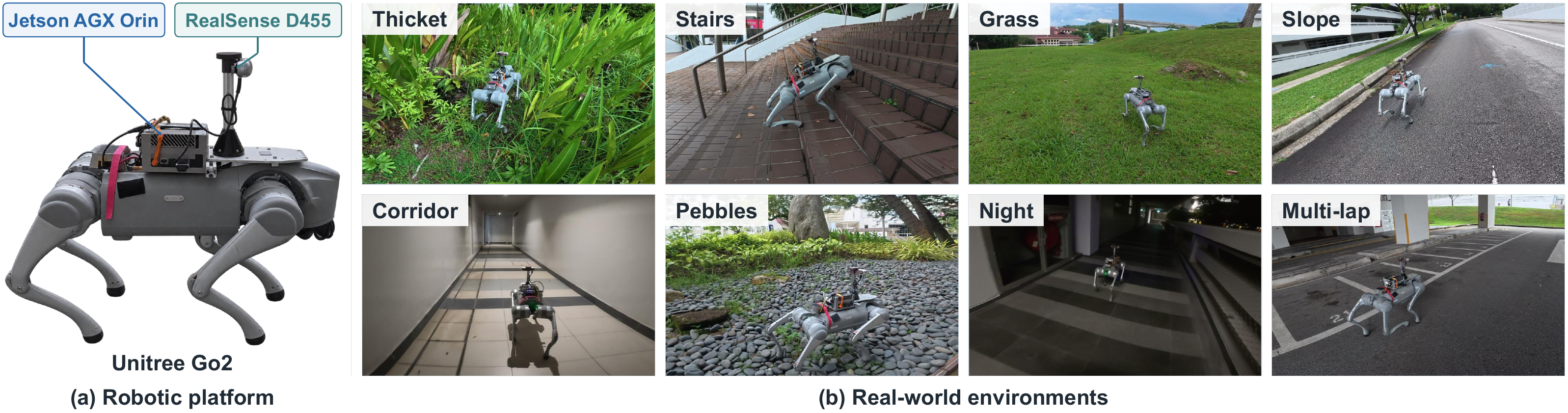}
  \caption{Low-cost quadruped platform and real-world test environments. (a) Unitree Go2 equipped with an Intel RealSense D455 camera and an NVIDIA Jetson AGX Orin. (b) Test scenes, including dense vegetation, stairs, grass, slopes, a textureless corridor, pebbles, a walkway at night, and a multi-lap route.}
  \label{fig:go2}
\end{figure*}

\input{figures/go2_table}

\subsubsection{Result Analysis}
Table~\ref{tab:sota} summarizes the results.
On the normal sequences, AirLeg achieves the best result on seven of the eight sequences and the second best on K\"{A}B-1.
Its average ATE of 0.15\,m is more than 25\% lower than those of the best competing methods, $\sqrt{\text{VINS}}$ and OKVIS2 (0.21\,m and 0.22\,m).
On the challenging sequences, AirLeg achieves the best result on five of the nine sequences and the second best on ETH-1. Its average ATE of 0.23\,m is much lower than that of the best competitor that completes all of them (IS, 2.08\,m).
The most extreme sequences cause failures for competitors of every sensor modality.
At night in SPX-3, three visual-inertial methods diverge, and three others are evaluated only on partial trajectories.
In CON-1, the camera stream drops out twice, for 13.9\,s and 15.2\,s, inside a dark and textureless building, and all visual and visual-inertial methods that finish the sequence drift by more than 9\,m.
In ICE-1, the robot stands still on the ice while it is pushed for about 27\,m, so that the leg-inertial methods observe no motion and drift by more than 4\,m.
In SBB-1, loose rail ballast with repetitive texture and strong overexposure degrades both modalities, and even the best competitor, OpenVINS, has an error of 0.82\,m, more than twice that of AirLeg (0.36\,m).
Overall, AirLeg is the only method that completes all 17 sequences with an error below 0.4\,m, whereas all competitors fail or exceed 4\,m at least once.

\textbf{Discussion.}
The results reveal a consistent pattern across the sensor configurations.
The leg-inertial methods never fail catastrophically, but they drift with distance and terrain, from about 0.5\% of the walked distance on the normal sequences to 1.4\%--7.3\% on slippery and deformable terrain.
The visual and visual-inertial methods are accurate when the images are informative, but each of them breaks down on some visually challenging sequences.
The existing visual-inertial-leg methods add handcrafted leg kinematics with fixed noise models, which prevents some visual failures (\eg Cerberus on SPX-3 and SBB-1, where VINS-Fusion diverges) but degrades the accuracy on the normal sequences (1.07\,m for Cerberus versus 0.80\,m for VINS-Fusion), because the leg velocity errors are fused as if they were reliable.
In contrast, AirLeg benefits from both modalities.
Its leg odometry constrains every keyframe and carries the estimate through visual gaps, such as the camera dropouts in CON-1, while the online VCE down-weights the leg factors when the feet slip or sink, and the online adaptation corrects the leg odometry on unseen terrain, as on SNOW-3.
On ICE-1 and RIV-1, however, AirLeg (0.31\,m and 0.29\,m) remains less accurate than the best competitors (0.02\,m and 0.13\,m), since the leg odometry is unreliable on ice even when the robot steps, and the leg-aided front end drifts in some intervals on soft sand.
We attribute the overall advantage mainly to the leg odometry that is adapted to the real robot instead of being modeled by hand, and to its fusion with vision.

\subsection{Real-World Tests on a Low-Cost Quadruped}\label{sec:exp-go2}

\begin{figure}[!t]
  \centering
  \setlength{\abovecaptionskip}{-0.4cm}
  \includegraphics[width=\columnwidth]{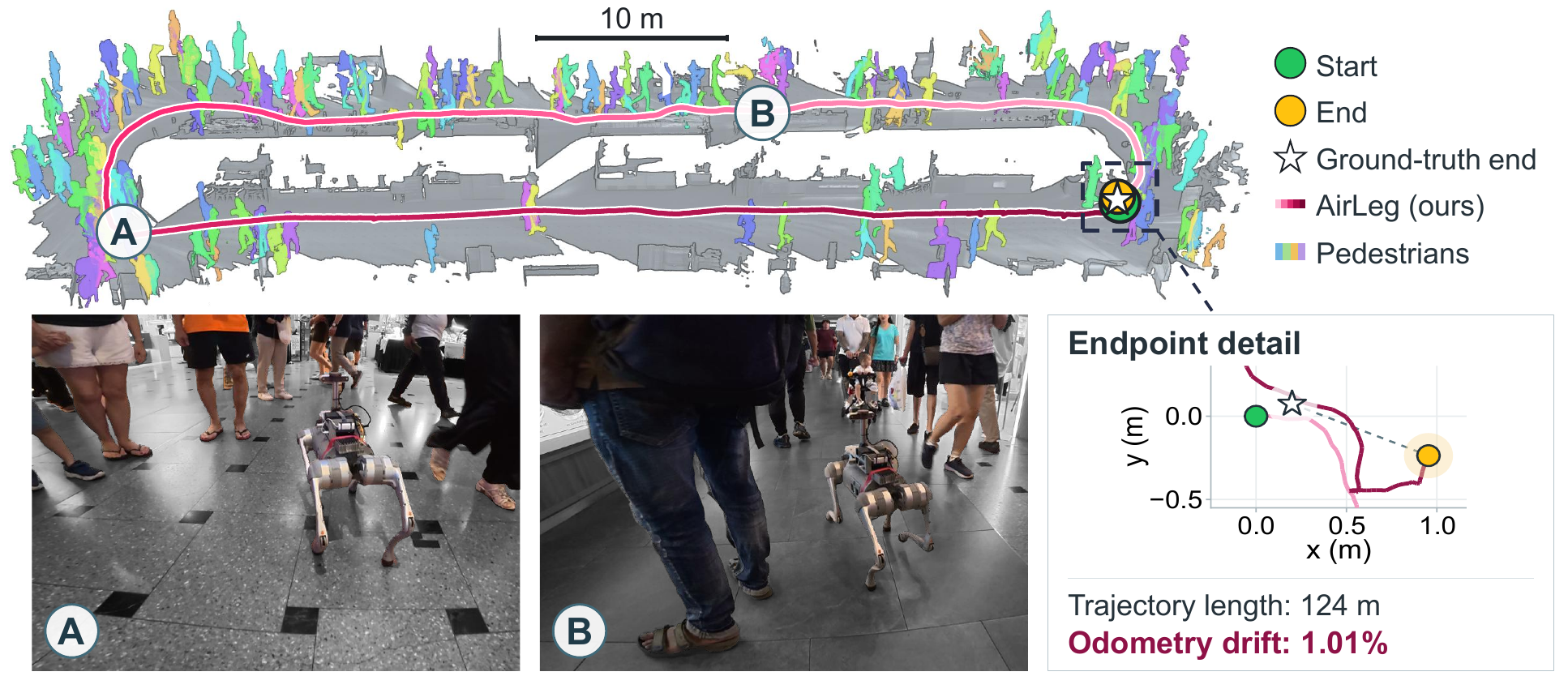}
 \caption{Test in a crowded shopping mall (Crowd in Table~\ref{tab:go2}). Top: map with the trajectory of AirLeg and the pedestrians. Bottom: views at A and B, and the trajectory near the start and end, with a drift of 1.01\% over 124\,m.}
  \label{fig:crowd}
\end{figure}

The GrandTour dataset is recorded by ANYmal~D, an industrial-grade quadruped with a high-quality sensor suite.
In practice, however, many legged robots are low-cost platforms with consumer-grade sensors.
To evaluate AirLeg in this setting, we build a data collection platform based on a Unitree Go2 quadruped equipped with an Intel RealSense D455 camera and an NVIDIA Jetson AGX Orin (Fig.~\ref{fig:go2}a).
The stereo images (640$\times$480) are recorded at 10\,Hz, and the body IMU and the joint encoders of the Go2 at 500\,Hz.
We record about 0.5\,h of training data for the self-supervised sim-to-real adaptation and ten test sequences with a total length of 1.8\,km.
The test sequences pose diverse challenges to both vision and leg odometry, including a dense thicket that repeatedly occludes the camera, stairs with pitch angles up to 32$^\circ$, flat grass, asphalt roads with slopes, a low-texture corridor with white walls, pebbles, a walkway at night, a route of five laps, a garden route with steps, and a crowded shopping mall with many moving pedestrians (Fig.~\ref{fig:go2}b, Fig.~\ref{fig:crowd}, and Table~\ref{tab:go2}).

For comparison, we select nine representative methods of Table~\ref{tab:sota} that run in real time and perform well in Section~\ref{sec:exp-sota}, covering leg-inertial, visual-inertial, and VIL estimators.
All competing methods use the raw 500\,Hz IMU and joint measurements, whereas AirLeg downsamples them to 125\,Hz as the input of its leg odometry network for a balance between efficiency and accuracy.
For the ground truth, we follow the UMA-VI dataset~\cite{zuniga2020vi} and let every test sequence start and end at the same place, in front of the same calibration board.
The start and end segments are reconstructed by SFM with the board, which provides the ground-truth relative pose of the left camera between the start and the end.
We therefore align each trajectory at the start and report the position error at the end in percent of the path length, where the estimated and ground-truth poses at the start and the end are each averaged over the period in which the robot stands still in front of the board.
As shown in Table~\ref{tab:go2}, AirLeg achieves an average drift of 0.59\%, less than 40\% of that of the best competing method, OKVIS2 (1.55\%), while three visual-inertial methods, Voxel-SVIO, $\sqrt{\text{VINS}}$, and OpenVINS, fail on the stairs.
It stays within about 1\% on all sequences and achieves the lowest drift on eight of the ten sequences, in particular in the low-texture corridor (0.55\%), where all other methods drift by more than 1.1\%, and in the crowded mall (1.01\%), where all other methods drift by at least 1.25\% or fail, even with the same pedestrian masks.

\input{figures/loop_table}

These results highlight the value of fusing vision, legs, and IMU on low-cost platforms.
First, we find that the visual-inertial methods depend on the quality of the IMU.
The body IMU of the Go2 is strongly affected by the vibrations of the legs: its noise while walking is much higher than at standstill.
As a result, the visual-inertial methods degrade significantly compared with the GrandTour dataset and end with mean rotation errors of 3.5$^\circ$ to 13.4$^\circ$, whereas AirLeg ends with 2.0$^\circ$.
Second, AirLeg is outperformed only on the two shortest sequences, thicket (56\,m) and pebbles (33\,m), where the differences to the best methods correspond to only 0.12\,m and 0.27\,m at the end.
Finally, all these results are obtained with only about 0.5\,h of training data recorded on the new robot without ground truth, which shows that the self-supervised adaptation transfers AirLeg to a new platform with little effort.

\subsection{Loop Closure and Global Refinement}\label{sec:exp-loop}

\begin{figure}[!t]
  \centering
  \setlength{\abovecaptionskip}{-0.4cm}
  \includegraphics[width=\columnwidth]{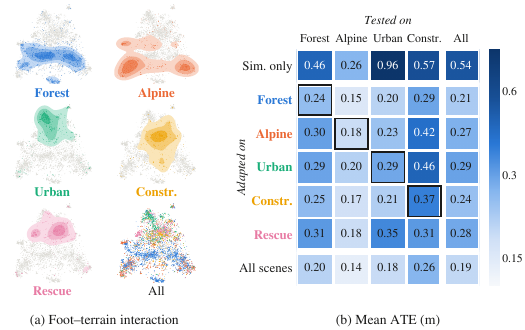}
  \caption{Generalization of the sim-to-real adaptation across scenes.
  (a) t-SNE embedding of the foot-terrain interaction of individual strides; each map highlights one scene category over all strides (gray).
  (b) Mean ATE of the networks adapted on different scenes (rows) and tested on each scene category (columns); the framed cells denote the same scene for adaptation and testing.}
  \label{fig:scene-generalization}
  \vspace{-0.8em}
\end{figure}

The comparison in Table~\ref{tab:sota} disables loop closure in all methods.
To evaluate the accuracy of the complete systems, we select the five test sequences that contain loops.
For the baselines, we select four methods that perform well in Table~\ref{tab:sota} and support loop closure, \ie two visual-inertial systems, OKVIS2 and ORB-SLAM3, and two VIL systems, Cerberus and GeoFlow-SLAM, and enable the loop closure of their released implementations.
With loop closure, we evaluate the trajectory after the final global optimization of each method.
Table~\ref{tab:loop} summarizes the results.
With the global refinement, AirLeg achieves the best result on four of the five sequences.
It reduces the error on all five sequences, \eg from 0.24\,m to 0.14\,m on LEE-1 and from 0.08\,m to 0.04\,m on SNOW-3, while the vertical error drops below 2\,cm on four of them.
The baselines benefit from their loop closure to different degrees.
ORB-SLAM3 and GeoFlow-SLAM detect very few loops in these sequences.
OKVIS2 benefits the most, \eg from 0.52\,m to 0.18\,m on LEE-1, and the final bundle adjustment even recovers SNOW-3, where its online estimate diverges, but it still fails on SPX-3.
Cerberus corrects loops on four sequences, but its error remains above 0.2\,m on all of them.

\subsection{Generalization across Scenes}\label{sec:exp-scene}

The network evaluated in Section~\ref{sec:exp-sota} is adapted on training sequences from all scenes.
In practice, a robot may be adapted in one environment and deployed in another.
To evaluate the generalization of our system in this setting, we group the training sequences into five scene categories, \ie forest trails, alpine sites, urban areas, construction and industrial sites, and a search-and-rescue training site, adapt one network on each category, and test all of them on the 17 test sequences, grouped into the same categories except search and rescue, which has no test sequence.
We visualize the distribution of the data from these scenes with t-SNE in Fig.~\ref{fig:scene-generalization}a, where each stride is embedded according to the contact forces and joint torques of the legs and the vibration measured by the IMU.
The scenes clearly differ in how the feet interact with the ground: the rigid ground of the urban, construction, and search-and-rescue scenes occupies one region, the soft forest soil a second one, and the snow and rock of the alpine scenes lie at its periphery.

Despite these differences, the adaptation transfers well across scenes (Fig.~\ref{fig:scene-generalization}b).
Every single-scene network reduces the mean ATE of the simulation-only network from 0.54\,m to 0.21--0.29\,m, recovering most of the gain of adapting on all scenes (0.19\,m).
Surprisingly, data from the deployment scene brings no clear advantage: the framed in-domain cells are never clearly better than the others in their column, and the network adapted on urban areas is even less accurate on the urban test sequences (0.29\,m) than the one adapted on forest trails (0.20\,m).
A possible explanation is that much of the adaptation corrects the discrepancy between the simulated and the real robot, which is present on every terrain, rather than terrain-specific characteristics.
What matters instead is the amount and the variety of the data.
The forest trails provide the largest set (80\,min), whose strides cover the widest part of the embedding in Fig.~\ref{fig:scene-generalization}a, and their network is the most accurate single-scene network in every column, whereas the networks adapted on the four smaller sets of 29--39\,min reach 0.24--0.29\,m, the least accurate being the one adapted on the smallest set, \ie the urban areas.
As the amount of data may play a key role, we next study how much real-world data the adaptation requires.

\subsection{Effect of Adaptation Data Duration}\label{sec:exp-data-duration}

\begin{figure}[!t]
  \centering
  \setlength{\abovecaptionskip}{-0.5cm}
  \includegraphics[width=\columnwidth]{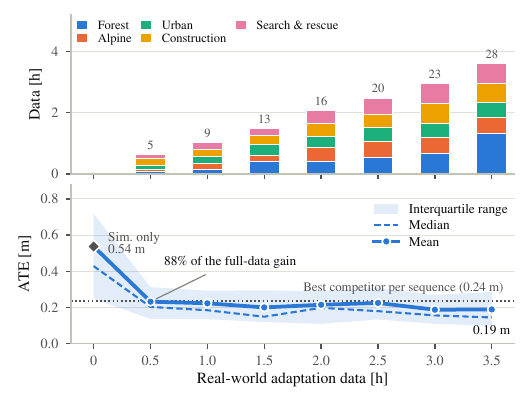}
  \caption{Effect of the amount of real-world adaptation data.
  Top: composition of the nested training sets by scene category, with the number of sequences above each bar.
  Bottom: ATE on the 17 test sequences of Table~\ref{tab:sota}.}
  \label{fig:data-duration}
\end{figure}

Although the sim-to-real adaptation needs no ground truth, it still needs real-world data of the target robot.
To study how much data is required, we repeat the adaptation on nested subsets of the 28 training sequences and evaluate each adapted network with the complete system on the 17 test sequences, both from the GrandTour dataset~\cite{tuna2026grandtour}.
Starting from all training sequences, which contain about 3.5\,h of data, we remove one sequence at a time from the scene category that contributes the most data, while keeping at least one sequence of each of the five categories (forest, alpine, urban, construction site, and search-and-rescue site).
This yields seven training sets of 5 to 28 sequences with about 0.5 to 3.5\,h of data in steps of 0.5\,h (Fig.~\ref{fig:data-duration}, top).
The 0\,h setting uses the network pre-trained in simulation (Section~\ref{sec:sim}) without adaptation.

Fig.~\ref{fig:data-duration} shows that the adaptation is data-efficient.
Without real-world data, the mean ATE is 0.54\,m, and the simulation-trained leg odometry causes large errors on sequences where the system relies on it, \eg 1.20\,m on ETH-1 and 1.03\,m on CON-1.
With only 0.5\,h of data from five sequences, the mean ATE drops to 0.23\,m, which recovers 88\% of the improvement obtained with all data and is already slightly lower than the mean of the best competing method on each sequence (0.24\,m).
At the same time, the mean vertical error decreases by about one third, from 0.11\,m to 0.07\,m.
With more data, the mean ATE decreases gradually from 0.23\,m with 0.5\,h to 0.19\,m with 3.5\,h, with small fluctuations in between.
These results show that a small amount of real-world training data without ground truth already suffices to achieve satisfactory accuracy, and that more data improves it further, although slowly.

\subsection{Ablation Study}\label{sec:exp-ablation}

\input{figures/ablation_s3}

To quantify the contribution of each component, we remove the components from the full system one at a time and evaluate each variant on the GrandTour dataset, averaging the results over two runs.
Table~\ref{tab:ablation-s3} shows that removing any component increases the error, most of all the sim-to-real adaptation (0.537\,m).
Among the others, the largest increase occurs without the long-window fusion: the leg odometry is then fused only in the short window of the front end, which cannot correct its slowly accumulating drift, and the ATE more than doubles, from 0.189\,m to 0.414\,m.
Within the long window, the long-span landmarks are also essential, as removing them raises the ATE to 0.310\,m.
Without supervision refinement, the network is adapted to the real robot only with the visual-only BA estimates as fixed pseudo labels, and the supervising map is not refined with the leg odometry.
This raises the ATE from 0.189\,m to 0.274\,m, and mostly on the challenging sequences (0.344\,m versus 0.225\,m), \eg from 0.077\,m to 0.515\,m on ALB-3 and from 0.286\,m to 0.565\,m on RIV-1, which shows that the errors of the fixed visual labels are transferred to the network and that jointly refining the supervision is necessary.
With the map refined but its gradient stopped, \ie without the indirect term in \eqref{eq:chain}, the ATE is 0.202\,m: most of the improvement comes from the refined map, and the indirect gradient provides the rest.
Removing the online adaptation, the foot-terrain constraint, or the online VCE also raises the ATE, to 0.205--0.211\,m, and mostly on the challenging sequences (0.252--0.261\,m versus 0.225\,m).
The contact offset averages about 4\,cm on mud (ALB-3) and 2\,cm on snow (SNOW-3) but stays within 2\,mm on urban ground, and fixing it to zero on the six sequences where it exceeds 1\,cm raises the vertical error from 0.069\,m to 0.077\,m.

The last two rows evaluate each modality alone.
Without any leg input, our system drifts on CON-1, so this sequence is excluded from its averages; on the other 16 sequences, its ATE is 0.311\,m, compared with 0.180\,m for the full system.
The gap is also large on the normal sequences (0.336\,m versus 0.148\,m), which indicates that the leg odometry improves the accuracy even when vision is reliable, rather than serving only as a fallback.
The leg odometry alone reaches 0.968\,m, lower than the leg-inertial methods in Table~\ref{tab:sota}, but its vertical error of 0.541\,m is ten times that of the full system.

\subsection{Efficiency Analysis}\label{sec:exp-runtime}

\begin{table}[t]
  \centering
  \caption{Runtime and resource usage analysis on a Jetson AGX Orin.}
  \label{tab:runtime}
  \footnotesize
  \setlength{\tabcolsep}{4pt}
  \begin{tabular}{@{}lccc@{}}
    \toprule
    \multirow{2}{*}{Module} & \multirow{2}{1.15cm}{\centering Trigger rate [Hz]} & \multicolumn{2}{c}{Runtime [ms]} \\
    \cmidrule(l){3-4}
    & & w/o adapt. & w/ adapt. \\
    \midrule
    \multicolumn{4}{@{}l}{\textit{Leg odometry forward}} \\
    \quad Network inference & 125 & 4.1 & 4.9 \\
    \quad Integration & 125 & 0.1 & 0.1 \\
    \multicolumn{4}{@{}l}{\textit{Frame pose estimation}} \\
    \quad Feature extraction & 10 & 16.9 & 19.3 \\
    \quad Feature matching & 10 & 14.8 & 16.5 \\
    \quad Pose estimation & 10 & 2.6 & 2.6 \\
    \multicolumn{4}{@{}l}{\textit{Short-window optimization}} \\
    \quad Stereo matching & 1.29 & 14.3 & 15.7 \\
    \quad Local BA & 1.29 & 51.3 & 51.6 \\
    \multicolumn{4}{@{}l}{\textit{Long-window fusion}} \\
    \quad Dense stereo of superframes & 0.90 & 78.7 & 89.6 \\
    \quad Optimization & 0.13 & 123 & 123 \\
    \multicolumn{4}{@{}l}{\textit{Leg odometry backward}} \\
    \quad Adapter training and validation & 0.11 & -- & 1922 \\
    \quad Adapter update & 0.08 & -- & 9.6 \\
    \midrule
    \multicolumn{2}{@{}l}{CPU usage [cores]} & 1.2 & 1.4 \\
    \multicolumn{2}{@{}l}{GPU memory [GB]} & 1.8 & 2.6 \\
    \bottomrule
  \end{tabular}
\end{table}

To assess the onboard efficiency, we run AirLeg on the Jetson AGX Orin of the Go2 platform in Section~\ref{sec:exp-go2} and evaluate it without and with the online adaptation on the 10\,Hz stereo images (Table~\ref{tab:runtime}).
In terms of runtime, the per-frame cost is dominated by the learned feature extraction and matching (31.7\,ms in total), whereas the more expensive local BA runs only at keyframes, \ie at about 1.3\,Hz.
The most expensive modules, \ie the dense stereo of superframes and the long-window optimization, run at low rates of 0.90\,Hz and 0.13\,Hz in separate threads, and thus hardly block the front end.
Similarly, the online adaptation runs in a separate process, where a round of training and validation takes about 1.9\,s and is triggered only about every 9\,s, so that the adaptation runs only at a low frequency.
Its cost lies mainly in GPU memory: it adds 0.8\,GB, mostly for the separate training process, but little CPU usage (1.4 versus 1.2 cores).
Since the training shares the GPU with the VO, the GPU modules become slightly slower, \eg dense stereo from 78.7\,ms to 89.6\,ms, but the end-to-end latency of non-keyframes increases by only 4\,ms, from 34\,ms to 38\,ms.
Overall, even with the online adaptation, AirLeg uses only 1.4 of the 12 CPU cores and 2.6\,GB of GPU memory, and runs in real time on the embedded computer of the low-cost Go2.

\section{Conclusion}\label{sec:conclusion}
In this paper, we presented AirLeg, a VIL SLAM system for legged robots whose leg odometry is adapted to the real robot without ground truth or additional sensors.
The leg odometry network is pre-trained in simulation, adapted to the real robot by a self-supervised imperative learning framework, in which the leg odometry helps refine the visual map through multi-sensor BA and the refined map in turn supervises the network, and further adapted online during operation.
In the online system, a dual-window estimator fuses the adapted leg odometry with vision and the IMU, a deformation-aware foot-terrain constraint limits the vertical drift, and variance component estimation adaptively weights the heterogeneous constraints.
On the GrandTour dataset, using only the built-in IMU of the robot, AirLeg achieves average ATEs of 0.15\,m and 0.23\,m on the normal and challenging sequences and is the only method that completes all 17 test sequences with an error below 0.4\,m.
On a low-cost Go2 quadruped, it achieves the lowest average drift with only about 0.5\,h of training data recorded without ground truth, and it runs in real time on an NVIDIA Jetson AGX Orin.
These results show that self-supervised learning of leg odometry, combined with a tight fusion of vision, legs, and IMU, enables accurate and robust localization of legged robots with low-cost sensors.


\ifCLASSOPTIONcaptionsoff
  \newpage
\fi

{
    \bibliographystyle{IEEEtran}
    \bibliography{papers}
}

{
\normalfont

\input{authors}
}

\vfill

\end{document}

%% file: figures/sota_table.tex
\providecommand{\camicon}{\raisebox{-0.25ex}{\includegraphics[height=2.4ex]{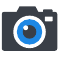}}}
\providecommand{\imuicon}{\raisebox{-0.25ex}{\includegraphics[height=2.4ex]{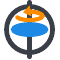}}}
\providecommand{\legicon}{\raisebox{-0.25ex}{\includegraphics[height=2.4ex]{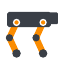}}}
\providecommand{\fail}{\textcolor{red}{F}}
\providecommand{\yr}[1]{{\scriptsize\color{black!55}#1}}
\begin{table*}[!t]
  \centering
  \caption{ATE RMSE [m] on the GrandTour dataset. Best and second-best results are shown in \textbf{bold} and \underline{underlined}.}
  \label{tab:sota}
  \vspace{-0.7em}
  \footnotesize
  \setlength{\tabcolsep}{2.6pt}
  \renewcommand{\arraystretch}{1.05}
  \ifdefined\sotaAbox\else\newsavebox{\sotaAbox}\fi
  \sbox{\sotaAbox}{%
  \begin{tabular}{@{}l @{\hspace{13pt}}c@{\hspace{13pt}} c@{\hspace{5.8pt}}c@{\hspace{5.8pt}}c@{\hspace{5.8pt}} C{1.16cm}C{1.16cm}C{1.16cm}C{1.16cm}C{1.16cm}C{1.16cm}C{1.16cm}C{1.16cm} c@{}}
    \multicolumn{14}{c}{(a) Normal sequences} \\
    \toprule
    \multirow{2}{*}{Method} & \multirow{2}{*}{Year} & \camicon & \imuicon & \legicon & PIL-2 & HEAP-1 & ROOT-1 & TRIM-1 & LEICA-1 & EIG-2 & ETH-2 & K\"{A}B-1 & \multirow{2}{*}{Avg.} \\
    & & V & I & L & 274\,m & 162\,m & 178\,m & 268\,m & 153\,m & 198\,m & 211\,m & 205\,m & \\
    \midrule
    MUSE~\cite{nistico2025muse} & \yr{2025} & \xx & \cc & \cc & 2.47 & 0.99 & 10.4 & 2.63 & 1.44 & 0.88 & 1.33 & 2.47 & 2.82 \\
    CAPO~\cite{sun2026capo} & \yr{2026} & \xx & \cc & \cc & 6.77 & 0.57 & 2.14 & 2.36 & 0.32 & 0.65 & 0.37 & 1.64 & 1.85 \\
    InEKF~\cite{hartley2020contact} & \yr{2020} & \xx & \cc & \cc & 1.19 & 0.52 & 1.20 & 1.16 & 0.92 & 0.74 & 1.42 & 1.78 & 1.12 \\
    IS~\cite{nistico2026proprioceptive} & \yr{2026} & \xx & \cc & \cc & 1.25 & 0.88 & 0.89 & 1.02 & 0.49 & 0.75 & 0.70 & 1.70 & 0.96 \\
    IEKF~\cite{nistico2026proprioceptive} & \yr{2026} & \xx & \cc & \cc & 1.25 & 0.88 & 0.88 & 1.02 & 0.49 & 0.74 & 0.70 & 1.69 & 0.96 \\
    \midrule
    MAC-VO~\cite{qiu2025macvo} & \yr{2025} & \cc & \xx & \xx & 1.86 & 1.30 & 3.23 & 2.59 & 1.18 & 0.51 & 2.43 & 4.73 & 2.23 \\
    DROID-SLAM~\cite{teed2021droid} & \yr{2021} & \cc & \xx & \xx & 0.26 & 0.23 & 0.61 & 0.47 & 0.36 & \underline{0.13} & 0.70 & 0.74 & 0.44 \\
    \midrule
    SchurVINS~\cite{fan2024schurvins} & \yr{2024} & \cc & \cc & \xx & 0.35 & 0.60 & 1.19 & 1.27 & 1.08 & 1.02 & 0.32 & 2.00 & 0.98 \\
    VINS-Fusion~\cite{qin2019general} & \yr{2019} & \cc & \cc & \xx & 0.60 & 0.49 & 1.66 & 0.73 & 0.99 & 0.22 & 0.39 & 1.33 & 0.80 \\
    OpenVINS~\cite{geneva2020openvins} & \yr{2020} & \cc & \cc & \xx & \underline{0.12} & \underline{0.10} & 0.28 & 0.35 & 0.94 & \fail & 1.91 & \textbf{0.17} & 0.55$^{\ast}$ \\
    Voxel-SVIO~\cite{yuan2025voxel} & \yr{2025} & \cc & \cc & \xx & 0.54 & 0.43 & 0.47 & 0.50 & \underline{0.24} & 0.19 & \underline{0.20} & 0.71 & 0.41 \\
    ORB-SLAM3~\cite{campos2021orb} & \yr{2021} & \cc & \cc & \xx & 0.18$^\dagger$ & 0.24 & 0.28 & 0.36$^\dagger$ & 0.30 & 0.22$^\dagger$ & 0.37 & 0.19$^\dagger$ & 0.27 \\
    OKVIS2~\cite{leutenegger2022okvis2} & \yr{2022} & \cc & \cc & \xx & 0.15 & 0.13 & 0.23 & 0.28 & 0.30 & 0.14 & \underline{0.20} & 0.33 & 0.22 \\
    $\sqrt{\text{VINS}}$~\cite{peng2025sqrtvins} & \yr{2025} & \cc & \cc & \xx & 0.18 & \textbf{0.09} & \underline{0.18} & \underline{0.23} & 0.27 & \underline{0.13} & 0.29 & 0.33 & \underline{0.21} \\
    \midrule
    Quad-Stack~\cite{aditya2025robust} & \yr{2025} & \cc & \cc & \cc & 0.97 & 2.07 & 22.4 & 4.39 & 10.2 & 2.31 & 13.1 & 13.3 & 8.58 \\
    Cerberus2.0~\cite{yang2025multi} & \yr{2025} & \cc & \cc & \cc & 8.71 & 2.18 & 3.17 & 2.46 & 2.66 & 2.89 & 2.10 & 4.99 & 3.64 \\
    Cerberus~\cite{yang2023cerberus} & \yr{2023} & \cc & \cc & \cc & 0.86 & 0.86 & 1.05 & 1.35 & 1.29 & 0.51 & 0.65 & 1.98 & 1.07 \\
    GeoFlow-SLAM~\cite{xiao2025geoflow} & \yr{2025} & \cc & \cc & \cc & 0.79 & 0.59 & 0.88 & 1.12 & 0.72 & 0.99 & 1.34 & 0.86 & 0.91 \\
    \textbf{AirLeg (Ours)} & \yr{--} & \cc & \cc & \cc & \textbf{0.09} & \textbf{0.09} & \textbf{0.14} & \textbf{0.18} & \textbf{0.14} & \textbf{0.12} & \textbf{0.10} & \underline{0.31} & \textbf{0.15} \\
    \bottomrule
  \end{tabular}}
  \usebox{\sotaAbox}

  \vspace{0.6em}
  \begin{tabular*}{\wd\sotaAbox}{@{}l @{\hspace{13pt}}c@{\hspace{13pt}} @{\extracolsep{\fill}}C{1.16cm}C{1.16cm}C{1.16cm}C{1.16cm}C{1.16cm}C{1.16cm}C{1.16cm}C{1.16cm}C{1.34cm} c@{}}
    \multicolumn{12}{c}{(b) Challenging sequences (V / L: challenging for vision / for leg odometry)} \\
    \toprule
    \multirow{4}{*}{Method} & \multirow{4}{*}{Year} & SPX-3 & ETH-1 & LEE-1 & CON-1 & ICE-1 & ALB-3 & SNOW-3 & RIV-1 & SBB-1 & \multirow{4}{*}{Avg.} \\
    & & 101\,m & 205\,m & 332\,m & 263\,m & 56\,m & 78\,m & 145\,m & 204\,m & 311\,m & \\
    & & \scriptsize Night, low texture & \scriptsize Rain, dynamics & \scriptsize Lighting change & \scriptsize Camera dropout & \scriptsize Slipping, pushed & \scriptsize Mud, sinkage & \scriptsize Sinkage, slipping & \scriptsize Soft sand, steep climb & \scriptsize Loose ballast, overexposure & \\
    & & V & V & V & V & L & L & L & L & V+L & \\
    \midrule
    MUSE~\cite{nistico2025muse} & \yr{2025} & 0.41 & 1.89 & 6.81 & 1.46 & 4.08 & 2.50 & 2.86 & 2.45 & 5.69 & 3.13 \\
    CAPO~\cite{sun2026capo} & \yr{2026} & \underline{0.13} & 0.45 & 0.96 & 1.00 & 4.12 & 3.48 & 0.79 & 7.18 & 2.10 & 2.25 \\
    InEKF~\cite{hartley2020contact} & \yr{2020} & 0.70 & 0.73 & 2.07 & 9.43 & 5.09 & 2.16 & 2.91 & 2.54 & 13.8 & 4.39 \\
    IS~\cite{nistico2026proprioceptive} & \yr{2026} & 0.19 & 0.57 & 1.15 & 0.97 & 4.11 & 1.95 & 3.06 & 2.85 & 3.91 & \underline{2.08} \\
    IEKF~\cite{nistico2026proprioceptive} & \yr{2026} & 0.19 & 0.57 & 1.14 & \underline{0.96} & 4.11 & 1.95 & 3.07 & 2.86 & 3.91 & 2.09 \\
    \midrule
    MAC-VO~\cite{qiu2025macvo} & \yr{2025} & 1.11 & 1.84 & 8.65 & 36.8 & 0.21 & 0.29 & 0.44 & 2.15 & 13.0 & 7.16 \\
    DROID-SLAM~\cite{teed2021droid} & \yr{2021} & 0.19 & 0.47 & 0.71 & 24.2 & \textbf{0.02} & \underline{0.05} & \textbf{0.08} & 1.67 & 2.19 & 3.29 \\
    \midrule
    SchurVINS~\cite{fan2024schurvins} & \yr{2024} & 0.25 & \fail & 0.53 & \fail & \fail & 0.96 & 0.25 & 0.71 & 2.64 & 0.89$^{\ast}$ \\
    VINS-Fusion~\cite{qin2019general} & \yr{2019} & \fail & 0.30 & 0.82 & 9.90 & 0.16 & 1.15 & 0.16 & 0.53 & \fail & 1.86$^{\ast}$ \\
    OpenVINS~\cite{geneva2020openvins} & \yr{2020} & \fail & \fail & 1.62 & 46.1 & 0.22 & \textbf{0.04} & 0.22 & \textbf{0.13} & \underline{0.82} & 7.03$^{\ast}$ \\
    Voxel-SVIO~\cite{yuan2025voxel} & \yr{2025} & 0.22 & 0.43 & \underline{0.39} & \fail & \underline{0.06} & 0.06 & 0.14 & 0.44 & 9.08 & 1.35$^{\ast}$ \\
    ORB-SLAM3~\cite{campos2021orb} & \yr{2021} & 4.61$^\dagger$ & 0.30$^\dagger$ & 0.56 & \fail & 0.08 & 0.07 & 0.15 & 0.23 & 0.85 & 0.86$^{\ast}$ \\
    OKVIS2~\cite{leutenegger2022okvis2} & \yr{2022} & \fail & \textbf{0.09} & 0.52 & \fail & 0.08 & 0.06 & \fail & \underline{0.17} & 1.32 & 0.37$^{\ast}$ \\
    $\sqrt{\text{VINS}}$~\cite{peng2025sqrtvins} & \yr{2025} & 0.30$^\dagger$ & 0.34$^\dagger$ & 0.38$^\dagger$ & \fail & 0.23 & \underline{0.05} & \underline{0.12} & \textbf{0.13} & 1.82 & 0.42$^{\ast}$ \\
    \midrule
    Quad-Stack~\cite{aditya2025robust} & \yr{2025} & 10.8 & 1.86 & 43.1 & 39.3 & 10.0 & 1.65 & 0.91 & -- & 4.89 & 14.1$^{\ast}$ \\
    Cerberus2.0~\cite{yang2025multi} & \yr{2025} & 1.04 & 3.07 & 10.7 & \fail & 4.17 & 4.19 & 6.97 & 9.47 & 7.95 & 5.95$^{\ast}$ \\
    Cerberus~\cite{yang2023cerberus} & \yr{2023} & 0.37 & 0.93 & 1.52 & \fail & 4.06 & 0.40 & 0.33 & 1.69 & 8.39 & 2.21$^{\ast}$ \\
    GeoFlow-SLAM~\cite{xiao2025geoflow} & \yr{2025} & 4.99$^\dagger$ & 1.34 & \fail & \fail & 0.38 & 0.28 & 0.36 & -- & 2.10 & 1.58$^{\ast}$ \\
    \textbf{AirLeg (Ours)} & \yr{--} & \textbf{0.10} & \underline{0.25} & \textbf{0.24} & \textbf{0.33} & 0.31 & 0.08 & \textbf{0.08} & 0.29 & \textbf{0.36} & \textbf{0.23} \\
    \bottomrule
  \end{tabular*}

  \vspace{0.3em}
  \parbox{\wd\sotaAbox}{\scriptsize\raggedright
    \fail: failure, \ie the method lost track or drifted excessively (ATE $\geq$ 100\,m or $\geq$ 20\% of the walked distance).\\
    $^\dagger$: late initialization (more than 10\,s after the start) or re-initialization, so only a partial trajectory was evaluated, and the entry is not ranked.\\
    --: not available, since the RGB-D images required by the method were not recorded in this sequence.\\
    $^{\ast}$: some sequences are excluded from the average.\par}
    \vspace{-1.3em}
\end{table*}

%% file: figures/go2_table.tex
\providecommand{\camicon}{\raisebox{-0.25ex}{\includegraphics[height=2.4ex]{figures/icons/camera.pdf}}}
\providecommand{\imuicon}{\raisebox{-0.25ex}{\includegraphics[height=2.4ex]{figures/icons/imu.pdf}}}
\providecommand{\legicon}{\raisebox{-0.25ex}{\includegraphics[height=2.4ex]{figures/icons/leg.pdf}}}
\providecommand{\fail}{\textcolor{red}{F}}
\begin{table*}[!t]
  \centering
  \caption{Drift error on the Go2 sequences. Best and second-best results are shown in \textbf{bold} and \underline{underlined}.}
  \label{tab:go2}
  \footnotesize
  \setlength{\tabcolsep}{3pt}
  \renewcommand{\arraystretch}{1.05}
  \begin{threeparttable}
  \begin{tabular}{@{}l ccc C{1.03cm}C{1.03cm}C{1.03cm}C{1.03cm}C{1.03cm}C{1.03cm}C{1.03cm}C{1.28cm}C{1.03cm}C{1.03cm} c@{}}
    \toprule
    \multirow{2}{*}{Method} & \camicon & \imuicon & \legicon & Thicket & Stairs & Grass & Slope & Corridor & Pebbles & Night & Multi-lap & Mixed & Crowd$^{\dagger}$ & \multirow{2}{*}{Avg.} \\
    & V & I & L & 56\,m & 139\,m & 157\,m & 335\,m & 92\,m & 33\,m & 165\,m & 521\,m & 181\,m & 124\,m & \\
    \midrule
    IEKF~\cite{nistico2026proprioceptive} & \xx & \cc & \cc & \underline{0.47\%} & 3.94\% & 2.07\% & 2.17\% & 2.38\% & 3.81\% & 1.53\% & 1.67\% & 3.90\% & 1.30\% & 2.32\% \\
    IS~\cite{nistico2026proprioceptive} & \xx & \cc & \cc & \textbf{0.43\%} & 3.91\% & 2.13\% & 1.88\% & 2.21\% & 3.75\% & \underline{1.40\%} & 1.71\% & 3.57\% & 1.50\% & 2.25\% \\
    CAPO~\cite{sun2026capo} & \xx & \cc & \cc & 0.63\% & 4.82\% & 1.23\% & \underline{0.97\%} & 2.64\% & 5.06\% & 1.69\% & 0.86\% & 3.04\% & 1.51\% & 2.25\% \\
    VINS-Fusion~\cite{qin2019general} & \cc & \cc & \xx & 0.80\% & 4.40\% & 1.35\% & 5.75\% & 3.34\% & 0.69\% & 4.84\% & 1.36\% & 3.05\% & 7.28\% & 3.29\% \\
    OpenVINS~\cite{geneva2020openvins} & \cc & \cc & \xx & 1.92\% & \fail & \underline{0.44\%} & 7.81\% & 2.46\% & \textbf{0.12\%} & 2.42\% & 3.04\% & 2.72\% & \underline{1.25\%} & 2.46\%$^{\ast}$ \\
    $\sqrt{\text{VINS}}$~\cite{peng2025sqrtvins} & \cc & \cc & \xx & 1.10\% & \fail & \fail & 2.79\% & \underline{1.14\%} & \underline{0.18\%} & 4.30\% & \underline{0.35\%} & 2.23\% & \fail & 1.73\%$^{\ast}$ \\
    Voxel-SVIO~\cite{yuan2025voxel} & \cc & \cc & \xx & 1.01\% & \fail & 0.89\% & 2.04\% & 2.18\% & 0.54\% & 2.67\% & 0.76\% & 3.56\% & 3.26\% & 1.88\%$^{\ast}$ \\
    OKVIS2~\cite{leutenegger2022okvis2} & \cc & \cc & \xx & 1.18\% & \underline{0.85\%} & 0.87\% & 2.78\% & 1.67\% & 0.53\% & 3.07\% & 0.90\% & \underline{1.37\%} & 2.26\% & \underline{1.55\%} \\
    Cerberus~\cite{yang2023cerberus} & \cc & \cc & \cc & 1.53\% & 5.11\% & 3.05\% & 9.69\% & 1.33\% & 4.68\% & 6.75\% & 3.10\% & 3.60\% & 8.15\% & 4.70\% \\
    \textbf{AirLeg (Ours)} & \cc & \cc & \cc & 0.64\% & \textbf{0.59\%} & \textbf{0.22\%} & \textbf{0.79\%} & \textbf{0.55\%} & 0.92\% & \textbf{0.44\%} & \textbf{0.21\%} & \textbf{0.53\%} & \textbf{1.01\%} & \textbf{0.59\%} \\
    \bottomrule
  \end{tabular}
  \begin{tablenotes}[flushleft]\scriptsize
    \item[] $^{\dagger}$: on the Crowd sequence, pedestrians and other dynamic objects are masked out in the images of all camera-based methods with the same YOLO11n-seg instance masks.
    \item[] \fail: failure (drift above 20\% of the path, or no pose over a large part of the sequence); $^{\ast}$: some sequences are excluded from the average.
  \end{tablenotes}
  \end{threeparttable}
  \vspace{-1.0em}
\end{table*}

%% file: figures/loop_table.tex
\providecommand{\fail}{\textcolor{red}{F}}
\begin{table}[!t]
  \centering
  \caption{ATE RMSE [m] without (w/o) and with (w/) loop closure. The best result of each sequence is shown in \textbf{bold}.}
  \label{tab:loop}
  \footnotesize
  \setlength{\tabcolsep}{2pt}
  \begin{tabular}{@{}llccccc@{\hspace{\dimexpr2\tabcolsep+4pt}}c@{}}
    \toprule
    Method & LC & ETH-2 & SPX-3 & LEE-1 & SNOW-3 & RIV-1 & Avg. \\
    & & 211\,m & 101\,m & 332\,m & 145\,m & 204\,m & \\
    \midrule
    OKVIS2~\cite{leutenegger2022okvis2} & w/o & 0.20 & \fail & 0.52 & \fail & 0.17 & 0.30$^{\ast}$ \\
     & w/ & 0.15 & \fail & 0.18 & 0.06 & \textbf{0.14} & 0.13$^{\ast}$ \\
    \midrule
    ORB-SLAM3~\cite{campos2021orb} & w/o & 0.37 & 4.61$^\dagger$ & 0.56 & 0.15 & 0.23 & 1.18 \\
     & w/ & 0.32 & 0.42$^\dagger$ & 0.64 & 0.16 & 0.20 & 0.35 \\
    \midrule
    Cerberus~\cite{yang2023cerberus} & w/o & 0.65 & 0.37 & 1.52 & 0.33 & 1.69 & 0.91 \\
     & w/ & 0.46 & 0.24 & 0.84 & 0.22 & 1.76 & 0.70 \\
    \midrule
    GeoFlow-SLAM~\cite{xiao2025geoflow} & w/o & 1.34 & 4.99$^\dagger$ & \fail & 0.36 & -- & 2.23$^{\ast}$ \\
     & w/ & 1.35 & 6.90 & \fail & 0.30 & -- & 2.85$^{\ast}$ \\
    \midrule
    \textbf{AirLeg (Ours)} & w/o & 0.10 & 0.10 & 0.24 & 0.08 & 0.29 & 0.16 \\
     & w/ & \textbf{0.09} & \textbf{0.09} & \textbf{0.14} & \textbf{0.04} & 0.22 & \textbf{0.12} \\
    \bottomrule
  \end{tabular}

  \vspace{0.3em}
  \parbox{\columnwidth}{\scriptsize\raggedright
    \fail: failure; $^\dagger$: partial trajectory; --: no RGB-D data; $^{\ast}$: average excluding \fail{} and --.\par}
    \vspace{-1.5em}
\end{table}

%% file: figures/ablation_s3.tex
\begin{table}[!t]
  \centering
  \caption{Ablation study on the GrandTour test sequences.}
  \label{tab:ablation-s3}
  \footnotesize
  \setlength{\tabcolsep}{1.5pt}
  \renewcommand{\arraystretch}{0.94}
  \begin{tabular*}{0.95\columnwidth}{@{\extracolsep{\fill}}lccc@{}}
    \toprule
    \multirow{2}{*}{Setting} & \multicolumn{3}{c}{ATE\,/\,vertical error [m]} \\
    \cmidrule(l){2-4}
    & Normal & Challenging & All \\
    \midrule
    \textbf{Full system} & \textbf{0.148\,/\,0.052} & \textbf{0.225\,/\,0.054} & \textbf{0.189\,/\,0.053} \\
    \midrule
    w/o indirect gradient & 0.156\,/\,0.054 & 0.243\,/\,0.058 & 0.202\,/\,0.056 \\
    w/o foot-terrain & 0.153\,/\,0.060 & 0.252\,/\,0.071 & 0.205\,/\,0.066 \\
    w/o online adaptation & 0.151\,/\,0.054 & 0.261\,/\,0.099 & 0.209\,/\,0.078 \\
    w/o VCE & 0.159\,/\,0.053 & 0.257\,/\,0.065 & 0.211\,/\,0.060 \\
    w/o supervision refinement & 0.195\,/\,0.060 & 0.344\,/\,0.117 & 0.274\,/\,0.090 \\
    w/o long-window landmarks & 0.264\,/\,0.057 & 0.351\,/\,0.071 & 0.310\,/\,0.064 \\
    w/o long-window fusion & 0.352\,/\,0.163 & 0.470\,/\,0.214 & 0.414\,/\,0.190 \\
    w/o sim-to-real adaptation & 0.476\,/\,0.063 & 0.592\,/\,0.145 & 0.537\,/\,0.106 \\
    \midrule
    Visual-inertial only & 0.336\,/\,0.079 & 0.285\,/\,0.055$^{\dagger}$ & 0.311\,/\,0.067$^{\dagger}$ \\
    Leg odometry only & 0.893\,/\,0.625 & 1.035\,/\,0.467 & 0.968\,/\,0.541 \\
    \bottomrule
  \end{tabular*}
  \par\vspace{0.2em}
  \begin{minipage}{0.95\columnwidth}
    \scriptsize $^{\dagger}$\,Mean without CON-1, on which the visual-inertial system drifts.
  \end{minipage}
  \vspace{-0.3em}
\end{table}